\pdfoutput=1

\documentclass[english,11pt]{article}
\usepackage[T1]{fontenc}
\usepackage[utf8]{inputenc}
\usepackage{geometry}
\usepackage{verbatim}
\usepackage{float}
\usepackage{bm}
\usepackage{amsmath}
\usepackage{amssymb}
\usepackage{amsfonts}
\usepackage{graphicx}
\usepackage{breakurl}
\usepackage{multirow}
\usepackage{mathtools}
\usepackage{makecell}
\usepackage{pgfplots}
\usepackage[htt]{hyphenat}
\usepackage{authblk}
\usepackage{listings}
\definecolor{dkgreen}{rgb}{0,0.6,0}
\definecolor{gray}{rgb}{0.5,0.5,0.5}
\definecolor{mauve}{rgb}{0.58,0,0.82}

\usepackage[most]{tcolorbox}

\tcbset{
  hwprompttemplate/.style={
    colback=white,
    colframe=blue!70!black,
    coltitle=white,
    fonttitle=\bfseries,
    colbacktitle=blue!80!black,
    title={Prompt Template for HW Problems},
    boxrule=1pt,
    arc=2pt,
    top=1mm,
    bottom=1mm,
    fontupper=\small,
    fontlower=\small
  },
    hwevaluationprompttemplate/.style={
    colback=white,
    colframe=blue!70!black,
    coltitle=white,
    fonttitle=\bfseries,
    colbacktitle=blue!80!black,
    title={Prompt Template for Evaluating the Responses},
    boxrule=1pt,
    arc=2pt,
    top=1mm,
    bottom=1mm,
    fontupper=\small,
    fontlower=\small
  },
  simoptprompttemplate/.style={
  colback=white,
    colframe=blue!70!black,
    coltitle=white,
    fonttitle=\bfseries,
    colbacktitle=blue!80!black,
    title={Prompt Template for SimOpt Problems},
    boxrule=1pt,
    arc=2pt,
    top=1mm,
    bottom=1mm,
    fontupper=\small,
    fontlower=\small
  }
}

\newtcolorbox{examplebox}[1][]{
  enhanced,
  breakable, 
  colback=pink!10,
  colframe=black,
  boxrule=1pt,
  arc=4pt,
  title={#1},
  fonttitle=\bfseries\color{white},
  colbacktitle=black,
  boxed title style={
    enhanced,
    colframe=black,
    boxrule=1pt,
    arc=4pt,
    top=1pt,bottom=1pt,left=6pt,right=6pt
  },
  attach boxed title to top left={xshift=12pt,yshift=-6pt},
}

\usepackage{tikz}
\usepackage{tikzsymbols}
\usetikzlibrary{arrows.meta, shapes.geometric, positioning, calc}
\tikzstyle{block1} = [rectangle, rounded corners, minimum width=3cm, minimum height=1cm, text centered, draw=black, fill=red!30]
\tikzstyle{block2} = [rectangle, rounded corners, minimum width=3cm, minimum height=1cm, text centered, text width=5cm, draw=black, fill=blue!20]
\tikzstyle{block3} = [rectangle, rounded corners, minimum width=3cm, minimum height=1cm, text centered, text width=5cm, draw=black, fill=orange!30]
\tikzstyle{block4} = [rectangle, rounded corners, minimum width=3cm, minimum height=1cm, text centered, draw=black, fill=green!30]
\tikzstyle{block5} = [rectangle, rounded corners, minimum width=3cm, minimum height=1cm, text centered, draw=black, fill=purple!30]
\tikzstyle{block6} = [rectangle, rounded corners, minimum width=1cm, minimum height=1.2cm, text centered, text width=2.5cm, draw=black, fill=red!30]
\tikzstyle{block7} = [rectangle, rounded corners, minimum width=1cm, minimum height=1.2cm, text centered, text width=4cm, draw=black, fill=blue!20]
\tikzstyle{block8} = [rectangle, rounded corners, minimum width=1cm, minimum height=1.2cm, text centered, text width=2.5cm, draw=black, fill=orange!30]
\tikzstyle{block9} = [rectangle, rounded corners, minimum width=1cm, minimum height=1.2cm, text centered, text width=2.5cm, draw=black, fill=green!30]
\tikzstyle{block10} = [rectangle, rounded corners, minimum width=1cm, minimum height=1.2cm, text centered, text width=3.8cm, draw=black, fill=purple!30]
\tikzstyle{arrow} = [thick,->,>=stealth]

\RequirePackage[authoryear]{natbib}
\RequirePackage[colorlinks,citecolor=blue,urlcolor=blue]{hyperref}

\usepackage[font=small]{caption}
\usepackage{cleveref}
\usepackage{lipsum}
\usepackage{subcaption}

\makeatletter
\def\blfootnote{\gdef\@thefnmark{}\@footnotetext}

\usepackage{amsthm}
\usepackage{dsfont}
\usepackage{array}
\usepackage{mathrsfs}

\usepackage{color}

\usepackage{enumitem}
\setlist[itemize]{leftmargin=1em}
\setlist[enumerate]{leftmargin=1em}
\allowdisplaybreaks

\usepackage[algo2e]{algorithm2e}
\usepackage{algorithmic}
\usepackage{arydshln}

\newcommand{\EE}{\mathbb{E}}

\newcommand{\NN}{\mathbb{N}}

\newcommand\myCref[1]{\@ifundefined{r@#1}{\Cref*{ec:#1}}{\Cref{#1}}}
\newcommand\myref[1]{\@ifundefined{r@#1}{\ref*{ec:#1}}{\ref{#1}}}

\makeatother

\theoremstyle{plain}  
\newtheorem{example}{\textbf{Example}}

 \theoremstyle{definition}
\newtheorem{remark}{\textbf{Remark}}

\numberwithin{equation}{section}
\numberwithin{theorem}{section}
\numberwithin{lemma}{section}
\numberwithin{corollary}{section}
\numberwithin{claim}{section}
\numberwithin{fact}{section}
\numberwithin{definition}{section}
\numberwithin{example}{section}
\numberwithin{assumption}{section}

\begin{document}

\title{Performance of LLMs on Stochastic Modeling Operations Research Problems: From Theory to Practice}

\author{Akshit Kumar}
\author{Tianyi Peng}
\author{Yuhang Wu}
\author{Assaf Zeevi}
\affil{Columbia Business School, Columbia University}

\date{This version: \today}

\maketitle

\begin{abstract}
Large language models (LLMs) have exhibited expert-level capabilities across various domains. However, their abilities to solve problems in Operations Research (OR)---the analysis and optimization of mathematical models derived from real-world problems or their verbal descriptions---remain underexplored. In this work, we take a first step toward evaluating LLMs' abilities to solve stochastic modeling problems, a core class of OR problems characterized by uncertainty and typically involving tools from probability, statistics, and stochastic processes. We manually procure a representative set of graduate-level homework and doctoral qualification-exam problems and test LLMs' abilities to solve them. We further leverage {\tt SimOpt}, an open-source library of simulation-optimization problems and solvers, to investigate LLMs' abilities to make real-world decisions under uncertainty. Our results show that, though a nontrivial amount of work is still needed to reliably automate the stochastic modeling pipeline in reality, state-of-the-art LLMs demonstrate proficiency on par with human experts in both classroom and practical settings. These findings highlight the potential of building AI agents that assist OR researchers and amplify the real-world impact of OR through automation.

\end{abstract}

\section{INTRODUCTION} \label{sec:intro}
Large language models (LLMs) have showcased stunning capabilities across a wide spectrum of tasks, from writing code and solving mathematical problems to understanding subtle context and simulating human behavior. These breakthroughs have sparked a wave of innovation in autonomous AI agents that can interact with dynamic environments and tackle complex decision-making tasks (\cite{shinn2023reflexion, schickToolformerLanguageModels2023, yao2023react}; see also \href{https://manus.im/}{Manus AI}, \href{www.cursor.com}{Cursor Agent}, and \href{https://www.anthropic.com/claude-code}{Claude Code}). As LLMs and agentic systems rapidly advance, the Operations Research (OR) community faces an exciting opportunity: to harness these tools for automating core workflows, accelerating discovery, and ultimately amplifying the reach and impact of OR in the real world.

\begin{figure}[h]
\begin{tcolorbox}
\begin{tikzpicture}[node distance=0.5cm]
\node (problem) [block6] {Real-world Problem};
\node (problemgiver) [block10, below = of problem] {Industry Stakeholders, Policy-makers, ...};
\node (abstractmodel) [block7, right = 1cm of problem] {Abstraction \& Modeling};
\node (analyzesimulate) [block8, right = of abstractmodel] {Analysis \& Optimization};
\node (testdeploy) [block9, right = of analyzesimulate] {Testing \& Deployment};
\draw [arrow] (problemgiver) -- (problem);
\draw [arrow] (problem) -- (abstractmodel);
\draw [arrow] (abstractmodel) -- (analyzesimulate);
\draw [arrow] (analyzesimulate) -- (testdeploy);
\draw[arrow](testdeploy.north)-- ++(0,+6mm)-- ([yshift=+6mm]analyzesimulate.north) -- (analyzesimulate.north); 
\draw[arrow](analyzesimulate.north)-- ++(0,+6mm)-- ([yshift=+6mm]abstractmodel.north) node[midway, above] {Iterate} -- (abstractmodel.north); 
\draw[arrow](abstractmodel.north)-- ++(0,+6mm)-- ([yshift=+6mm]problem.north) -- (problem.north); 

\node[block10, below= of analyzesimulate] (researcher) {OR Researcher}; (researcher) {OR Researcher};
\coordinate[above=2mm of researcher.north] (aux);
\draw[arrow]  (researcher) -- (aux) -| (abstractmodel);
\draw[arrow]  (aux) -| (analyzesimulate);
\draw[arrow]  (aux) -| (testdeploy);

\draw[<->, thick] (problemgiver.east) to node[midway, below right, align=center] {\hspace{-3em} Collaboration\\\hspace{-3em}\& Feedback} (researcher.west);
\end{tikzpicture}
\end{tcolorbox}
\caption{The standard pipeline of OR. Automating this pipeline with AI agents may significantly enhance the power of OR and amplify its real-world impact.}
\label{fig:orpipeline}
\end{figure}
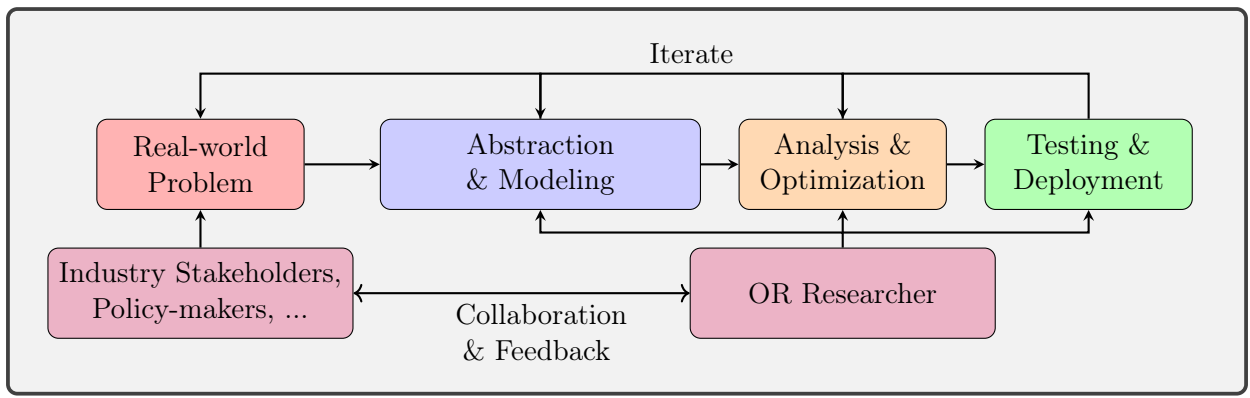

As a field, OR is fundamentally concerned with decision-making through mathematical modeling. After all, the real world is full of complexity, but by encoding that complexity into mathematical form, we gain a powerful language that liberates us from ad hoc reasoning and enables rigorous analysis and computational approaches for solving real-world problems. A typical OR pipeline begins with identifying and abstracting real-world challenges---often encountered by industry stakeholders and policy-makers---into a mathematical representation capturing essential data, objectives, and constraints. This representation undergoes analysis, simulation, and optimization, to yield actionable insights and viable solutions, subject to iterative refinement and practical validation prior to deployment (see Figure \ref{fig:orpipeline}). This comprehensive, principled approach has effectively addressed numerous complex real-world issues, including those in supply chain management, transportation, finance, retail, healthcare, and online platforms. A strong endorsement of this pipeline’s success is the INFORMS Franz Edelman Award, which have delivered over \$419 billion in cumulative benefits \citep{informs_edelman}.

But this pipeline does not come without cost. The OR workflow often requires researchers to continuously refine models, carefully balancing analytical tractability with practical relevance, with the guiding principle that all models are wrong, but some are more useful than others. This iterative process demands profound domain expertise and technical proficiency. After identifying a solution, researchers need to work with the stakeholders for testing and deployment and possibly circle back to earlier stages to revise and improve the model. All these steps are highly non-trivial, time-consuming, and can be both an art and a science. The whole procedure can take years. Therefore, we believe developing specialized OR agents that can assist researchers to automate this pipeline could unlock significant productivity gains and transformative impacts.

\subsection{Deterministic Optimization vs Stochastic Modeling}
\begin{figure}[h]
\begin{tcolorbox}
\begin{minipage}{0.5\textwidth}
\centering
\begin{tikzpicture}[node distance=2cm]
\node (decision) [block1] {Decision variable $x$};
\node (objandconst) [block2, below of=decision] {Closed-form objective and constraints $f(x)$, $g(x)$};
\node (optimizer) [block3, below of=objandconst] {Optimizer (e.g., Gurobi, cvxpy)};
\node (output) [block4, below of=optimizer] {Optimal solution $x^*$};
\draw [arrow] (decision) -- (objandconst);
\draw [arrow] (objandconst) -- (optimizer);
\draw [arrow] (optimizer) -- (output);
\node[above,font=\large\bfseries,yshift=1em] at (current bounding box.north) {Deterministic Optimization};
\end{tikzpicture}
\end{minipage}
\begin{minipage}{0.5\textwidth}
\centering
\begin{tikzpicture}[node distance=2cm]
\node (decision) [block1] {Decision variable $x$};
\node (simulator) [block5, below of=decision] {Simulator};
\node (objandconst) [block2, below of=simulator] {Estimated objective and constraints $\hat{f}(x)$ and $\hat{g}(x)$};
\node (optimizer) [block3, below of=objandconst] {Optimizer (e.g., Random search, Nelder-Mead)};
\draw [arrow] (decision) -- (simulator);
\draw [arrow] (simulator) -- (objandconst);
\draw [arrow] (objandconst) -- (optimizer);
\draw [arrow] (optimizer.east) -- ++ (1em,0) |-node [above] {iterate}  (decision);
\node[above,font=\large\bfseries,yshift=1em] at (current bounding box.north) {Simulation-Optimization};
\end{tikzpicture}
\end{minipage}
\end{tcolorbox}
\caption{Comparison of conventional deterministic optimization with simulation-optimization. In deterministic setting, the objective $f(x)$ and the constraints $g(x)$ are known closed-form functions, enabling a direct call to an optimizer. In simulation-optimization, the performance of a decision $x$ is observed through a stochastic simulator, producing noisy estimates of the objective and constraints $\hat{f}(x)$ and $\hat{g}(x)$ that must be re-sampled until estimation error is acceptable. The optimizer then usually selects a new candidate decision variable based on the estimated performance of the current decision variable. This process therefore operates in a loop, trading off additional replications (to reduce estimation error) and searching unexplored promising regions.}
\label{fig:comparison}
\end{figure}
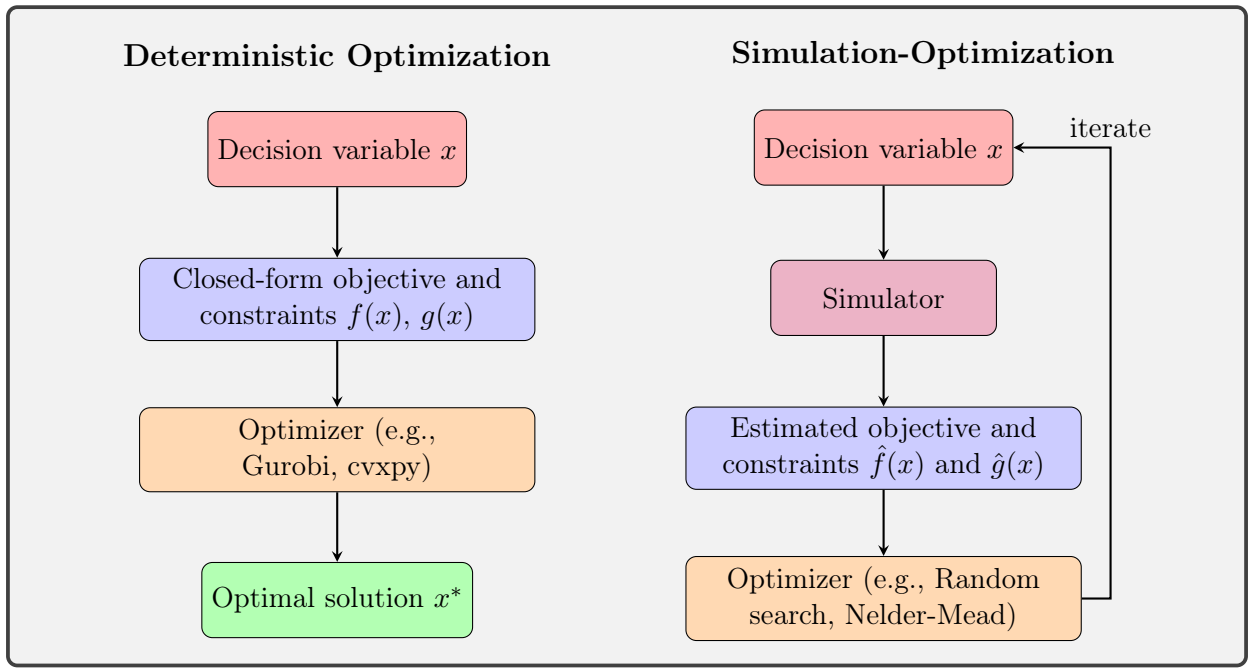

Recently, the OR community has begun to explore this direction with an emphasis on automating deterministic optimization problems \citep{ramamonjisonAugmentingOperationsResearch2022,ahmaditeshniziOptiMUS03UsingLarge2024,huangORLMCustomizableFramework2025,zhang2025or}. As a central pillar of OR, optimization problems have a wide range of real-world applications, and the ability to properly formulate and solve them is a key skill for OR practitioners. In contrast, the evaluation of LLMs' ability to handle \textit{stochastic modeling} problems, another pillar of OR, has been mostly absent in the literature. Stochastic modeling problems cast the decision-making process in a framework of uncertainty, which is inherent to many practical problems. Compared to optimization in a deterministic scenario, decision-making with uncertainty is intrinsically harder because a ``good'' decision has to work for a range of possible outcomes that are often unpredictable and may only be tractable en masse through an often unknown probability distribution. Consequently, stochastic modeling employs fundamentally different analytical and numerical methods---such as stochastic processes and simulation-based optimization---distinct from deterministic optimization methods (e.g., convex analysis and gradient-based algorithms). See Figure \ref{fig:comparison} for a comparison between deterministic optimization and \textit{simulation-optimization}.\footnote{A branch of OR known as robust optimization \citep{bertsimas2011theory} explicitly incorporates uncertainty into otherwise deterministic optimization formulations. Evaluating LLMs on their ability to understand and apply such techniques is also an interesting direction, which we leave for future exploration.} 

As an example, consider the Chess Matchmaking problem (detailed in Section \ref{sec:chessmm} and Appendix \ref{sec:simoptmoredetails}). Players, whose Elo scores\footnote{A score that measures the relative skill levels of players.} follow some probability distribution, arrive at the platform according to some random process, and two players are matched if their Elo difference is below some pre-determined threshold (i.e., the policy), while the average waiting time of the players should not exceed a specified threshold. This may seem to be a straightforward online matching problem, but there is \textit{no known analytical solution}, and one must resort to numerical solutions, where the choice of simulation-optimization solver is crucial. As illustrated by this example, the complexity of a stochastic modeling problem scales up quickly as multiple sources of randomness become entangled, and a real-world problem can easily become overly challenging. Another perhaps more well-known class of complex stochastic modeling problems are stochastic processing networks \citep{daiProcessingNetworksFluid2020}, notoriously challenging and often lacking general analytical solutions. Developing robust, general-purpose numerical solutions remains a significant research challenge. Therefore, in parallel to the effort of automating the deterministic optimization pipeline, we believe the automation of the stochastic modeling pipeline is equally important and introduces unique challenges.

\subsection{This work}
In this work, we take the first step towards building OR agents for assisting the stochastic modeling pipeline by focusing on moving from ``Abstraction \& Modeling'' to the ``Analysis \& Optimization'' step in Figure \ref{fig:orpipeline}. In particular, we aim to assess LLMs' capacities to solve formulated stochastic modeling problems in a real-world context, addressing both theoretical comprehension and practical numerical execution (see examples in the subsequent sections). We view the ability to solve a formulated problem in the stochastic modeling context as a core skill for assisting OR researchers and is a necessary step towards automating the entire pipeline.

We also recognize that supporting the transition from an industry-level ``Real-world Problem'' to the appropriate ``Abstraction \& Modeling,'' as well as facilitating the iterative collaboration between OR researchers and industry stakeholders, presents a unique challenging task for LLMs insofar as solving OR problems. Addressing this may require substantial research effort, and even designing a rigorous evaluation is highly non-trivial. While we view this as a promising direction for future work, in this paper we focus our attention on the ``Analysis \& Optimization'' stage, where evaluation is more tractable.

Specifically, we evaluate LLM performance on two classes of stochastic modeling problems:
\begin{itemize}
\item \textbf{Homework and exam problems.} These problems are designed to educate and assess students' understanding of how to analyze a model. Students are expected to reason about the model's theoretical properties and grasp the characteristics of a policy that can achieve desirable performance.
\item \textbf{Simulation-optimization problems.} These problems are meant to mimic real-world problems that often require numerical methods. Students are expected to design and implement an algorithm to simulate the model and optimize policy decisions accordingly.
\end{itemize}
Our key contributions and findings are as follows:
\begin{enumerate}
\item We construct a dataset of graduate-level stochastic modeling problems to test LLMs' abilities to solve them. We find strong performance overall with more open-ended modeling problems presenting a greater challenge.
\item We procure a set of qualification-exam problems and manually grade LLMs' answers to them. Our results show that they demonstrate comparable performance to human PhD candidates in the field.
\item Using the \href{https://github.com/simopt-admin/simopt}{{\tt SimOpt}} library, we show that top-performing LLMs can match the best in-house solvers on a range of simulation-optimization problems, underscoring their potential. However, our results also indicate that off-the-shelf use of LLMs is not yet reliable for automating the full stochastic modeling pipeline, suggesting the need for further adaptation and integration.
\end{enumerate}
In summary, our findings demonstrate that LLMs hold significant promise across stochastic modeling tasks---from analytical problem solving to simulation optimization---but realizing reliable end-to-end automation will require targeted refinement beyond off-the-shelf use. We hope this work encourages further research on automating stochastic modeling and contributes to building intelligent agents capable of autonomously understanding, modeling, and solving complex real-world problems.

The rest of the paper is organized as follows. In Section \ref{sec:litrev}, we review the related literature. In Section \ref{sec:hw}, we evaluate LLMs' performance on the homework problems dataset. In Section \ref{sec:qual}, we assess LLMs' performance on the qualifying exam problems. In Section \ref{sec:simopt}, we discuss the simulation-optimization problems and compare LLMs' solutions with standard solvers. Finally, in Section \ref{sec:disc}, we conclude the paper and discuss future directions.

\section{LITERATURE REVIEW} \label{sec:litrev}

LLMs have long been tested for their ability to solve math problems. GPT-3 was one of the first large-scale generative models to demonstrate strong zero-shot and few-shot capabilities across many tasks, including simple mathematical problems \citep{brownLanguageModelsAre2020}. Soon after, \cite{hendrycksMeasuringMathematicalProblem2021} introduced the MATH dataset, which consists of over 12,500 competition-level math problems covering algebra, geometry, number theory, and more, emulating the difficulty of middle and high-school math competitions. Concurrently, \cite{cobbeTrainingVerifiersSolve2021} released the GSM8K dataset that contains over 8,000 short-answer math word problems (grade-school level) focusing on arithmetic and multi-step reasoning. They also proposed a ``verifier'' model trained to check correctness of solutions. MATH and GSM8K datasets became widely used benchmarks to test reasoning and multi-step solution correctness. Many later papers used them as primary testbed, including the renowned chain-of-thought paper \citep{weiChainofThoughtPromptingElicits2023} and subsequent papers on LLM reasoning. Clearly, these math-exam-style problems cannot adequately test LLMs' abilities to solve complex modeling problems.

One direction that is more specialized and more challenging is applying LLMs to the stricter setting of formal logic and automated theorem proving (ATP). Unlike natural language math solutions, where partial correctness is sometimes acceptable, formal theorem proving requires exact logical derivations. Small mistakes are not tolerated by the proof checker. Early works on neural theorem proving started almost a decade ago \citep{rocktaschelEndtoEndDifferentiableProving2017,kaliszykHolStepMachineLearning2017,bansalHOListEnvironmentMachine2019,yangLearningProveTheorems2019} and typical theorem proving environments include Lean, Coq, Isabelle, and Mizar. They laid the groundwork for using general-purpose LLMs for theorem proving \citep{poluGenerativeLanguageModeling2020,zhengMiniF2FCrosssystemBenchmark2022,jiangThorWieldingHammers2022,yangLeanDojoTheoremProving2023}. Many challenges remain in this direction. For example, high-quality proof corpora are relatively small compared to internet-scale text. Theorem proofs can also be very long and even large LMs can hallucinate or lose track of intermediate states. See \cite{glazerFrontierMathBenchmarkEvaluating2024} for an example of scenarios where LLMs can still struggle. High-quality ATP abilities also do not translate directly to reliably solving modeling problems, where (open-ended) theoretical analyses are not the end goal but tools to design a policy that can achieve desirable performance in practice.

Another direction equips LLMs with code-writing to enhance their problem-solving abilities. The LLM's job is to write correct, logically consistent code, and then a downstream system (e.g., a Python interpreter) executes that code to obtain the result. \cite{gaoPALProgramaidedLanguage2023} is a good example, where the LLM's chain-of-thought is effectively replaced by or augmented with a Python function that solves each sub-step of the problem. \cite{romera-paredesMathematicalDiscoveriesProgram2024} is another prime example, where an LLM paired with a systematic evaluator pushed beyond the boundary of human knowledge on the cap set problem and discovered an asymptotic lower bound that was the largest improvement in 20 years. More broadly, this stream of literature is also related to code generation \citep{chenEvaluatingLargeLanguage2021,liCompetitionLevelCodeGeneration2022} and tool-use \citep{schickToolformerLanguageModels2023} by LLMs. Our work is broadly related to the vast literature on LLM-based AI agents as well (\cite{shinn2023reflexion, schickToolformerLanguageModels2023, wang2023voyager, yao2023react}; \href{https://manus.im/}{Manus AI}). More specifically, what we envision resembles a ``research agent'' that can facilitate human scientists make new discoveries (e.g., \href{https://sakana.ai/ai-scientist/}{Sakana AI}).

As mentioned before, the rapid development of LLMs' abilities to solve math problems generally does not consider the aspect of modeling. A series of work, primarily by OR researchers, have discussed successes in marrying LLMs with techniques of formulating optimization problems and solving them \citep{ramamonjisonAugmentingOperationsResearch2022,liSynthesizingMixedintegerLinear2023,astorgaAutoformulationMathematicalOptimization2024,xiaoCHAINOFEXPERTSWHENLLMS2024,mostajabdavehOptimizationModelingVerification2024,ahmaditeshniziOptiMUS03UsingLarge2024,huangORLMCustomizableFramework2025,jiangLLMOPTLearningDefine2025,zhang2025or}; valuable datasets and benchmarks have also been published. In contrast, studies on LLMs' abilities to solve stochastic modeling problems have been missing. With this paper, we aim to take the first step towards this direction and hope to inspire future research.

\section{HOMEWORK PROBLEMS TEST CASE} \label{sec:hw}
\subsection{Dataset}
To the best of our knowledge, no dataset featuring a list of stochastic modeling problems and solutions is publicly available. Ideally, the dataset should consist of a large number of mathematical models and analytical results from academic papers on stochastic modeling, but as a start we will focus on course settings as a prerequisite. Most textbooks available in digital forms do not have readily available solutions, and there are also copyright issues. Therefore, we manually sourced problems and solutions from related courses at our institution. 

The first version of the dataset has 175 problems and solutions, divided into three categories: probability, stochastic processes, and stochastic modeling. This categorization reflects both conceptual boundaries and a natural progression of study. The probability category establishes the theoretical foundation, covering topics such as measure-theoretic probability, convergence of random variables, the laws of large numbers, the central limit theorem, and large deviations. Building on this, the stochastic processes category introduces dynamic models including random walks, stopping times, martingales, Markov chains, and renewal and regenerative processes. Finally, the stochastic modeling category focuses on applications of these processes to real-world systems, with an emphasis on stochastic stability and queueing theory.

However, a quick inspection shows that a proportion of the problems are educative and not suitable for evaluation; see Example \ref{eg:easyhw1} and \ref{eg:easyhw2} for two sample problems. 

\begin{examplebox}[A classical result in probability theory]
\begin{example}\label{eg:easyhw1}
Let $X$ be a non-negative random variable with cumulative distribution function $F$. Show that $\mathbb{E}[X] = \int_0^\infty \bar{F}(x)dx$, where $\bar{F}(x) = 1 - F(x)$. 
\end{example}
\end{examplebox}

\begin{examplebox}[A classical result in stochastic processes]
\begin{example}\label{eg:easyhw2}
Prove the backwards martingale convergence theorem.
\end{example}
\end{examplebox}
We consider these problems to be ``classic'' and a quick test shows that LLMs can solve them very well, likely because their solutions appear frequently on the Internet. To make our evaluation more challenging and meaningful, we exclude these problems from consideration. The final dataset contains 71 problems and solutions, where 37 are on probability theory, 23 are on stochastic processes, and 11 are on stochastic modeling. Below are some examples:

\begin{examplebox}[A sample problem on probability]
\begin{example}\label{eg:datasetprob}
Consider a sequence of i.i.d. random variables $X_{1}, X_{2}, \ldots$, each having an exponential distribution with parameter 1. Let $M_{n}:=\max \left\{X_{1}, \ldots, X_{n}\right\}$. \\
(a) Let $Y_{n}=X_{n} \mathbf{1}\left\{X_{n} \leq \log n\right\}$ denote a truncated exponential, i.e., $Y_{n}=X_{n}$ if $X_{n} \leq \log n$ and equals zero otherwise. Prove that $Y_{n} \neq X_{n}$, i.o., almost surely. \\
(b) Prove that $M_{n} / \log n \rightarrow 1$ almost surely as $n \rightarrow \infty$, where $\log$ denotes the natural logarithm.
\end{example}
\end{examplebox}

\begin{examplebox}[A sample problem on stochastic processes]
\begin{example}\label{eg:datasetstochproc}
Consider an irreducible discrete-time Markov chain $X_n$ on a finite state space. Let $p$ denote its transition probability matrix. A function $f$ is said to be superharmonic if $f(x) \geq \sum_{y} p(x, y) f(y)$ or equivalently $f\left(X_{n}\right)$ is a supermartingale. Show that the Markov chain is recurrent if and only if every nonnegative superharmonic function is constant.
\end{example}
\end{examplebox}

\begin{examplebox}[A sample stochastic modeling problem]
\begin{example}\label{eg:datasetmodel}
Consider an inventory model in which demand for a commodity arrives at the end of each day. Successive demands are i.i.d. with distribution function $F(\cdot)$. The following $(s, S)$ policy is used: if the inventory level at the beginning of a day is less than or equal to $s$, we order up to $S$, and if the level is greater than $s$ no action is taken. Orders are assumed to filled instantaneously. Let $\left\{X_{n}\right\}_{n \geq 1}$ be the inventory level at the beginning of the $n$th day, right after delivery of inventory (if any). Suppose that $X_{1}=S$.\\
(a) Prove that $X_{n}$ is regenerative with points of regeneration that are given by the order epochs.\\
(b) Let $\left\{\tau_{n}\right\}$ denote the lengths of the regenerative cycles, derive an expression for $\mathbb{E} \tau_{1}$.\\
(c) Derive an expression for $\lim _{n \rightarrow \infty} \mathbb{P}\left(X_{n} \geq k\right)$.
\end{example}
\end{examplebox}

Naturally, these problems vary in difficulty and the knowledge required. Overall, we believe this dataset is representative of the problems that students in a graduate-level stochastic modeling course would encounter. See Appendix \ref{sec:example} for these example questions' solutions, LLMs' answers, and the grading process.

\subsection{LLMs, Prompt Template, and Solving the Problems}
We consider 6 LLMs: GPT-4o, o1, and o3-mini from \href{https://openai.com/}{OpenAI}; Claude 3.5 Sonnet from \href{https://www.anthropic.com/}{Anthropic} (calling the latest model Claude 3.7 Sonnet constantly generates server overload error); Llama 3.3 70B Instruct Turbo from \href{https://www.llama.com/}{Meta}; DeepSeek-R1 from \href{https://www.deepseek.com/}{DeepSeek}. While many other LLMs exist and may be of interest, we believe the ones selected are representative of popular and state-of-the-art LLMs. 
We use \href{https://platform.openai.com/docs/overview}{OpenAI API} for OpenAI models, \href{https://www.anthropic.com/api}{Claude API} for Anthropic models, and \href{https://www.together.ai/}{Together AI} for Llama and DeepSeek models. For each LLM and each problem, we call the corresponding API to obtain a solution by the following prompt template:

\begin{tcolorbox}[hwprompttemplate, breakable]
    You are given a problem from probability theory and stochastic modeling. You need to solve this problem rigorously to the best of your ability. Please provide as many details and explanations as you can. The problem is as follows: 
    \vspace{1em}

    \{{\tt problem}\}
\end{tcolorbox}

We use this prompt to evaluate the inherent capabilities of LLMs without additional enhancements. Current LLMs (e.g., o1, o3-mini, DeepSeek-R1) already incorporate reasoning abilities and advanced prompting and sampling techniques. Therefore, we intentionally avoid using complex prompts, chain-of-thought methods \citep{weiChainofThoughtPromptingElicits2023}, or specialized sampling approaches in our testing. While we acknowledge that fine-tuning and prompt engineering could potentially improve performance, we reserve these approaches for future exploration.

\subsection{Evaluating LLMs' Solutions}
Ideally, for a rigorous evaluation of LLMs' solutions, we would have qualified human graders to grade each of LLMs' answers. However, due to budget constraint, a more scalable and valid alternative is the ``LLM-as-a-judge'' approach \citep{zhengJudgingLLMasaJudgeMTBench2023}, where strong LLM models are used to evaluate LLM performance. In our case, we use GPT-4o as the judge. For each problem-answer pair, we obtain an evaluation from the judge by the following prompt template:

\begin{tcolorbox}[hwevaluationprompttemplate, breakable]
You are given a problem in probability theory and stochastic modeling along with its solution. You are also given a solution submitted by a student studying these topics. Your task is to evaluate the student's solution based on correctness and completeness. The score should be a number between 0 and 100, where 0 means completely incorrect and incomplete, and 100 means completely correct and complete. You should also provide detailed reasons for the score you assign. For example, you should indicate that some points are taken off because the student fails to use a key result or the student's reasoning is not rigorous enough. The problem is: 
\vspace{1em}

\{{\tt problem}\} 
\vspace{1em}

The correct solution is: 
\vspace{1em}

\{{\tt solution}\} 
\vspace{1em}

The student's solution is: 
\vspace{1em}

\{{\tt student\_solution}\} 
\vspace{1em}

Please provide the final score at the beginning of your response in double brackets, e.g., [[50]].
\end{tcolorbox}

By manually inspecting the judge's scores and comments, we find the evaluations reasonable and sometimes more comprehensive than those that would be provided by a human counterpart (see Section \ref{sec:example} for examples). To make the evaluation process more robust, we sample three independent scores from the judge (by calling the API three times) for each problem and take the average for the final score. This procedure mimics pooling scores assigned by three independent human graders. We found that, in general, the three scores do not differ much, underlining the reliability of using an LLM as the judge. In Section \ref{sec:qual}, we also compare the LLM's scores with human scores and show further evidence for alignment.

\subsection{Results}
The LLMs' average scores (and standard errors in parenthesis) are summarized in Table \ref{tab:hwproblems}.

\begin{table}[ht]
    \centering
    \begin{tabular}{ccccc}
    \hline
    \textbf{LLM} & \textbf{Prob theory} & \textbf{Stoch process} & \textbf{Stoch modeling} & \textbf{Total} \\
    \hline
    GPT-4o & 81.85 (1.95) & 78.72 (2.65) & 74.67 (1.57) & 79.72 (1.39) \\
    \textbf{o1} & \textbf{95.46 (0.47)} & \textbf{95.28 (0.56)} & \textbf{90.61 (1.96)} & \textbf{94.65 (0.48)} \\
    \textbf{o3-mini} & \textbf{96.97 (0.36)} & \textbf{96.25 (0.54)} & \textbf{92.58 (1.29)} & \textbf{96.05 (0.37)} \\
    Claude 3.5 Sonnet & 88.74 (1.09) & 86.96 (1.33) & 73.48 (3.74) & 85.80 (1.12) \\
    Llama 3.3 70B & 78.36 (2.47) & 75.99 (2.70) & 58.18 (3.49) & 74.46 (1.85) \\
    DeepSeek-R1 & 84.30 (2.00) & 73.33 (4.44) & 64.85 (5.03) & 77.73 (2.13) \\
    \hline
    \end{tabular}
    \caption{LLMs' scores on homework problems.}
    \label{tab:hwproblems}
\end{table}
While we believe that these scores are generally reasonable, we caution against taking these scores to be exact because they are, after all, generated by GPT-4o and thus may have certain bias \citep{zhengJudgingLLMasaJudgeMTBench2023}. Nevertheless, several qualitative observations can be made:
\begin{enumerate}
    \item First and foremost, if we set 60\% as the cutoff for passing a stochastic modeling course, all models would pass easily, with o1 and o3-mini being the top performing.
    \item Secondly, models with higher scores seem to have smaller standard errors as well. This may be because that they have been optimized to be more accurate and be more aligned with human preferences, which may reduce the variability in their answers.
    \item Lastly, for all models, stochastic modeling problems seem to be the hardest, perhaps due to their more open-ended nature. 
\end{enumerate}

\section{QUALIFICATION EXAM PROBLEMS} \label{sec:qual}

In light of the observations from the last section, we believe the state-of-the-art LLMs have abilities on par with human PhD students in the field. To make the evaluation more rigorous, we next procure a set of qualification-exam problems to test the LLMs. It is a common tradition in many fields to test a PhD student's ability with a qualification exam after the student has passed all required courses. Only those who pass the exam will be allowed to proceed to the next stage of their PhD and become PhD candidates. Since the LLMs exhibit superior performance in a course-like setting, it would be interesting to see how they perform in a real exam. To ensure fairness, we will manually grade the LLMs' answers; we will also compare our scores with those generated by GPT-4o to provide evidence for the validity of using GPT-4o as a grader.

Results from the last section suggest that o1 and o3-mini have the best performance. Therefore, we will manually grade the answers generated by these two models. As a comparison, we will also consider answers from Claude 3.5 Sonnet since it was the second runner-up. Recall that we observed that stochastic modeling problems seem to be the most challenging for these models. By manually inspecting the LLMs' answers, we are also convinced that the LLMs can answer any probability theory and stochastic process problems well in the exam. Therefore, we will focus on testing these LLMs on stochastic modeling problems. 

\subsection{Dataset}
We carefully select 8 stochastic modeling problems that have appeared in past qualification exams in the Decision, Risk, and Operations Division at Columbia Business School. A common theme of these problems is to model some real-life problem with an abstract model. The students are asked to analyze the model to arrive at sensible decisions for the real-life problem. Examples include modeling the spread of a virus by a branching process, variants of the newsvendor problem, optimizing assortment planning using Markov chains, and stability conditions for complex queueing systems. In other words, though these problems are still limited to clean models, they have a practical footing and are representative of ``analyzable'' real-life stochastic modeling problems. Since they are exam problems, they are also less open-ended compared to their counterparts in homework to ensure fair grading. On average, these problems are designed to be challenging for junior PhD students. Every year, a selected committee of faculty would design new qualification exam problems. Since past problems and their solutions are private, it is unlikely that LLMs' training set contains the exact problems. 

Due to departmental regulations, we are unable to share actual problems from past qualification exams. To provide a sense of the exam's format and style, we present below a sample stochastic modeling problem that is made available to first-year PhD students as part of their preparation. Since its primary purpose is to illustrate the structure and expectations of the exam, the problem is on the easier end of the spectrum compared to typical qualification exam problems. See Appendix \ref{sec:examplequalprob} for the problem's solution, an LLM's answer, and the grading.

\begin{examplebox}[A sample qualification exam stochastic modeling problem]
\begin{example}
Consider a non-preemptive FIFO queue with infinite buffer. Requests arrive according to a Poisson process with rate $\lambda$, and each has i.i.d. workload $w \sim \text{Exp}(\mu)$. The service proceeds as follows. Each request is initially processed for up to $\theta$ time units. If completed within $\theta$, it exits the system and the next request (if any) begins service. If not, then the system restarts service in a mode that is divided into two steps: 1) the request is broken into $n$ sub-tasks that are executed in parallel by $n$ servers, where the processing times of the sub-tasks are i.i.d., uniformly distributed in the interval $[0.5 w / n, 1.5 w / n]$; and 2) the results of all $n$ sub-tasks are combined to complete the service of the original request, in a step that can only commence after all $n$ sub-tasks are completed and its duration is exponentially distributed with rate $2 \mu$, independent of the processing times of the sub-tasks and of the processing requirements of any other requests.\\
(a) What is the stability condition for the system?\\
(b) What is the steady-state expected sojourn time for a new request? The sojourn time includes the waiting time in the queue and the service time.
\end{example}
\end{examplebox}

\textit{We want to emphasize that the qualification exams at Columbia Business School, especially the stochastic modeling part, are not easy by any means.} The core course that prepares students for this exam is widely known to be hard in the community. Students often spend a significant amount of time taking this class and preparing for this exam. Therefore, though we use a small dataset, the evaluation remains meaningful: as educators and researchers in OR, we are genuinely curious about how LLMs perform on these problems, and the results may inform how we teach and assess students in the future.

\subsection{Results}
We obtain answers from the LLMs in the same fashion as before. We also obtain GPT-4o's grading for these answers. The LLMs' scores (average scores and standard error in parenthesis where applicable) are summarized in Table \ref{tab:qualproblems}. As evidenced by the scores, though stochastic modeling problems seem to be more challenging for LLMs than other types of problems, with 60\% being the cutoff for passing, \textit{all three models can pass the qualification exam with flying colors}. As a reference, though exact score distributions of past qualification exams at Columbia Business School are unknown, across the years scores border-lining the 60\% cutoff or lower consistently occurred. In this sense, under the same evaluation framework for humans, these models have demonstrated capabilities at least on par with PhD candidates in the field. 

\begin{table}[ht]
    \centering
    \caption{LLMs' scores on qualification exam problems.}
    \begin{tabular}{ccccccc}
    \hline
    \textbf{Problem} & \multicolumn{2}{c}{\textbf{o1}} & \multicolumn{2}{c}{\textbf{o3-mini}} & \multicolumn{2}{c}{\textbf{Claude 3.5 Sonnet}} \\
    \hline
    Scored by & GPT-4o & human & GPT-4o & human & GPT-4o & human \\ 
    \hline 
    \textbf{Problem 1} & 95 & 94 & 93.33 & 94 & 81 & 74 \\
    \textbf{Problem 2} & 95 & 96 & 96.67 & 96 & 85 & 88 \\
    \textbf{Problem 3} & 95 & 91 & 95 & 94 & 86.67 & 85 \\
    \textbf{Problem 4} & 98.33 & 100 & 100 & 100 & 95 & 88 \\
    \textbf{Problem 5} & 96.67 & 95 & 96.67 & 100 & 91.67 & 100 \\
    \textbf{Problem 6} & 95 & 95 & 98.33 & 95 & 78.33 & 85 \\
    \textbf{Problem 7} & 94.67 & 85 & 98.33 & 87 & 85 & 80 \\
    \textbf{Problem 8} & 96.67 & 100 & 93.33 & 100 & 73.33 & 77 \\
    \hline
    \textbf{Total} & 95.79 (0.43) & 94.5 (1.61) & 96.46 (0.80) & 95.75 (1.47) & 84.5 (2.31) & 84.63 (2.66) \\
    \hline
    \end{tabular}
    \label{tab:qualproblems}
\end{table}

How well the top LLMs are solving these problems are indeed impressive! But of course, the LLMs' answers are not without flaws. Upon closely examining their answers, we found that, for almost all problems, they have the right intuition and high-level arguments. Where points are taken off, it was mostly because of missing calculations or proof steps. This finding suggests that these LLMs can be helpful assistants for solving stochastic modeling problems, but they tend to generate answers that might lack rigor.

\subsection{Grading Alignment}
\begin{figure}[ht]
\centering
\begin{subfigure}{.45\textwidth}
  \centering
  \includegraphics[width=1\linewidth]{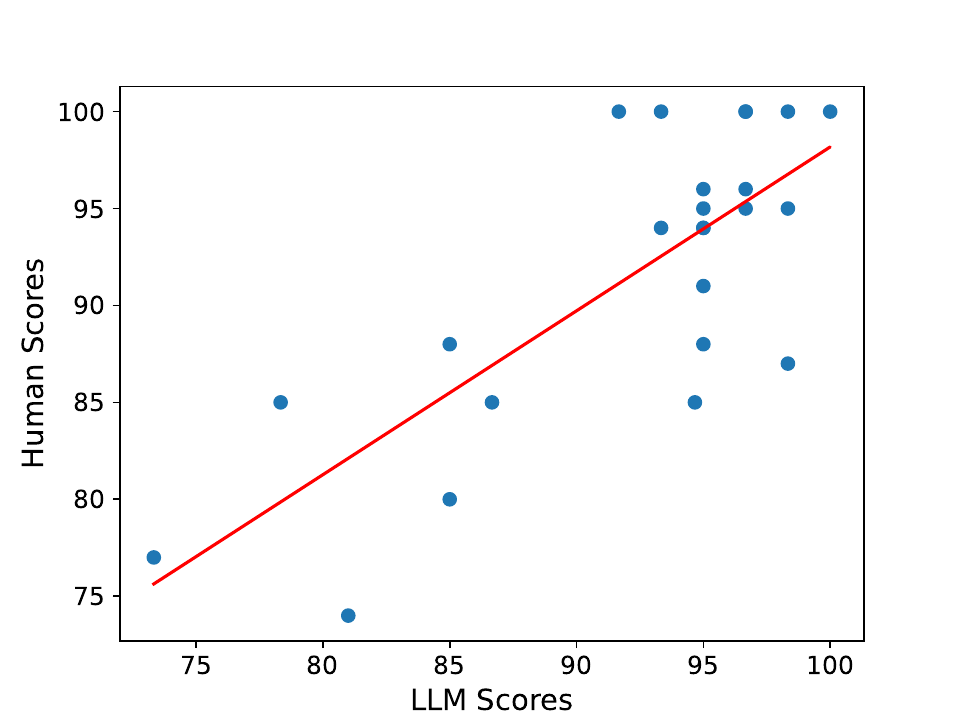}
  \caption{Scores given by GPT-4o vs. given by human.}
\end{subfigure}%
\begin{subfigure}{.45\textwidth}
  \centering
  \includegraphics[width=1\linewidth]{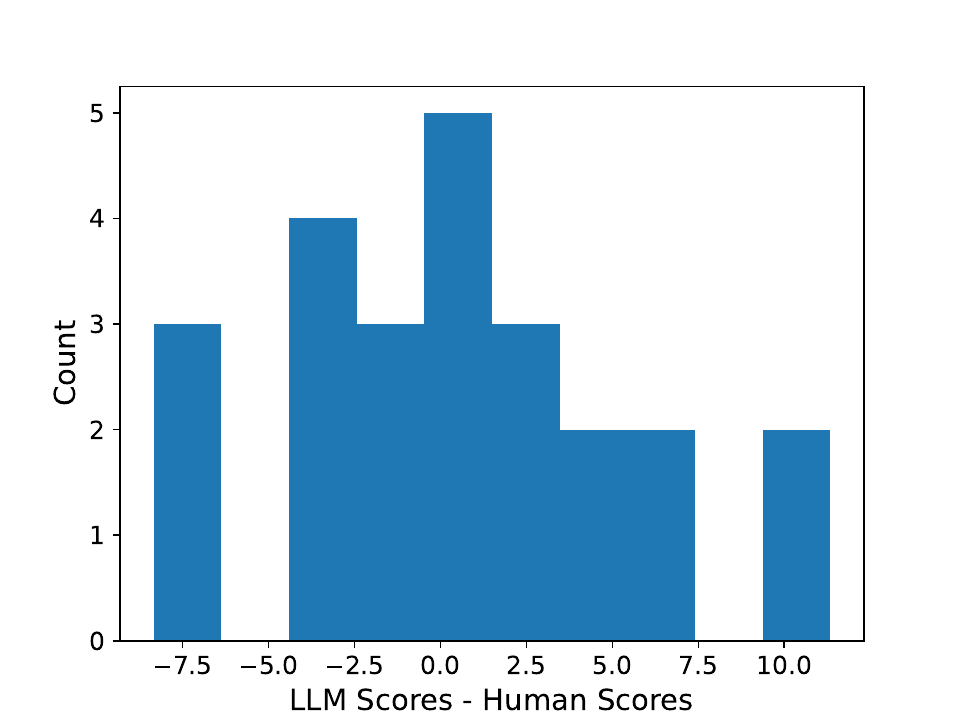}
  \caption{Distribution of grading differences. }
\end{subfigure}
\caption{Grading alignment.}
\label{fig:alignment}
\end{figure}

Given scores by humans, it is natural to ask whether the scores given by GPT-4o aligns well with ``real'' scores. Across the 24 scores, we calculated the Pearson correlation coefficient to be 0.77 (Figure \ref{fig:alignment}, left panel). The distribution of score differences (GPT-4o's scores minus human's scores) is approximately symmetric with a mean of 0.63 and standard error of 1.00 (Figure \ref{fig:alignment}, right panel). The absolute difference has a mean of 3.82 and a maximum of 11.33. Since the maximum score possible is 100, we believe the alignment between GPT-4o's grading and human's is well and the LLM-as-a-judge method is reliable in our context. We also acknowledge that a more comprehensive and large-scale evaluation would be beneficial and leave it for the future work.

\section{SIMULATION-OPTIMIZATION PROBLEMS} \label{sec:simopt}
Having shown that LLMs perform well in a classroom setting, in this section we discuss the ability of different LLMs in solving and implementing simulation-optimization methods on a testbed of problems that more closely resemble those encountered in practice. The code are available at \url{https://github.com/AkshitKumar/simopt-llm-evals}.

\subsection{Dataset and Evaluation}
We evaluate five LLMs---GPT-4o, o1, o3-mini, Claude 3.5 Sonnet, and DeepSeek-R1---on six optimization problems from the {\tt SimOpt} library, a well-known benchmark suite for noisy simulation optimization in OR \citep{simoptgithub}. Each model is prompted five times with problem descriptions from {\tt SimOpt}. We execute the Python code generated by the models and compare their solutions to those produced by baseline algorithms implemented in {\tt SimOpt}, including RandomSearch, ASTRO-DF, Nelder-Mead, STRONG, SPSA, ADAM, and ALOE (see \cite{dongEmpiricallyComparingFinitetime2017} for algorithmic details and comparisons). For each problem, we report the objective values achieved by the models and benchmark them against the best solver identified by \cite{dongEmpiricallyComparingFinitetime2017}.
These values may differ considerably due to variations in the simulation environments implemented by the different models. Consequently, we also focus on the algorithmic strategy that each model employs by manually analyzing the solutions they propose. Each model solution is allocated the same computational budget, where the budget refers to the number of simulation replications over the entire course of the search for optimal solutions \citep{simoptgithub}. To ensure consistency, we use the same prompt template for all five LLMs. The prompt is crafted by merging the problem descriptions from the {\tt SimOpt} library with a prompt template, which is then refined using o1 and manually verified for correctness. See Appendix \ref{sec:simoptmoredetails} for an example of the prompt and LLM's solution for the Chess Matchmaking problem. 

\subsection{Results}
\begin{figure}[ht]
    \centering
    \begin{subfigure}{0.48\textwidth}
        \includegraphics[width = \linewidth]{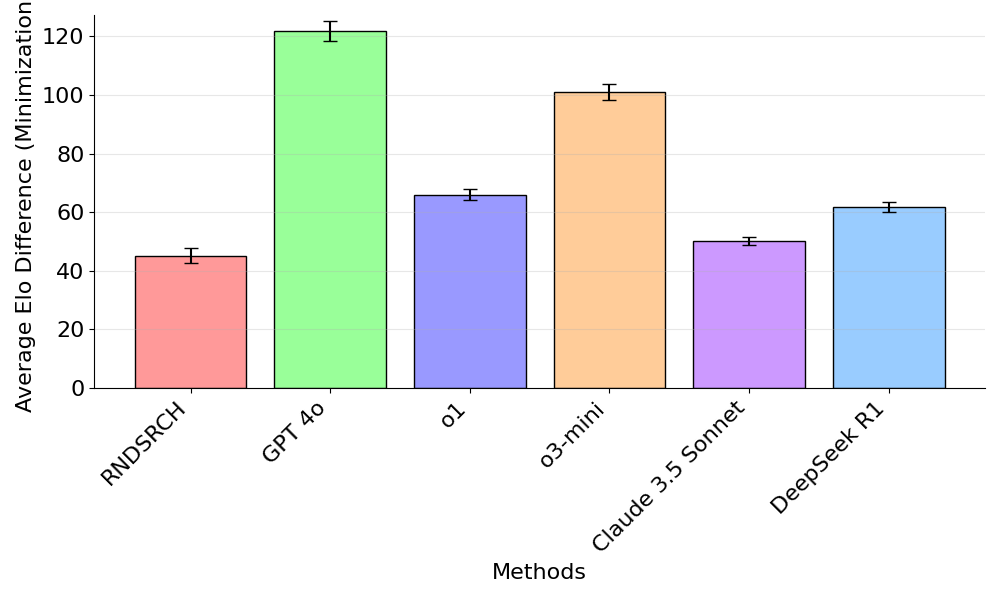}
        \caption{Chess Matchmaking}
    \end{subfigure}
    \begin{subfigure}{0.48\textwidth}
        \includegraphics[width = \linewidth]{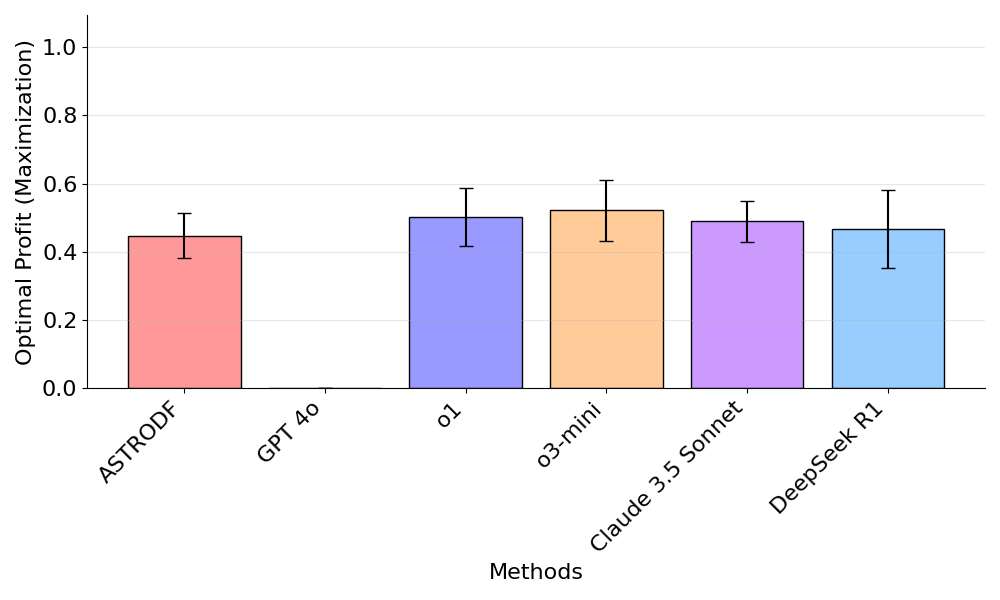}
    \caption{Continuous Newsvendor}
    \end{subfigure}
    \begin{subfigure}{0.48\textwidth}
    \includegraphics[width = \linewidth]{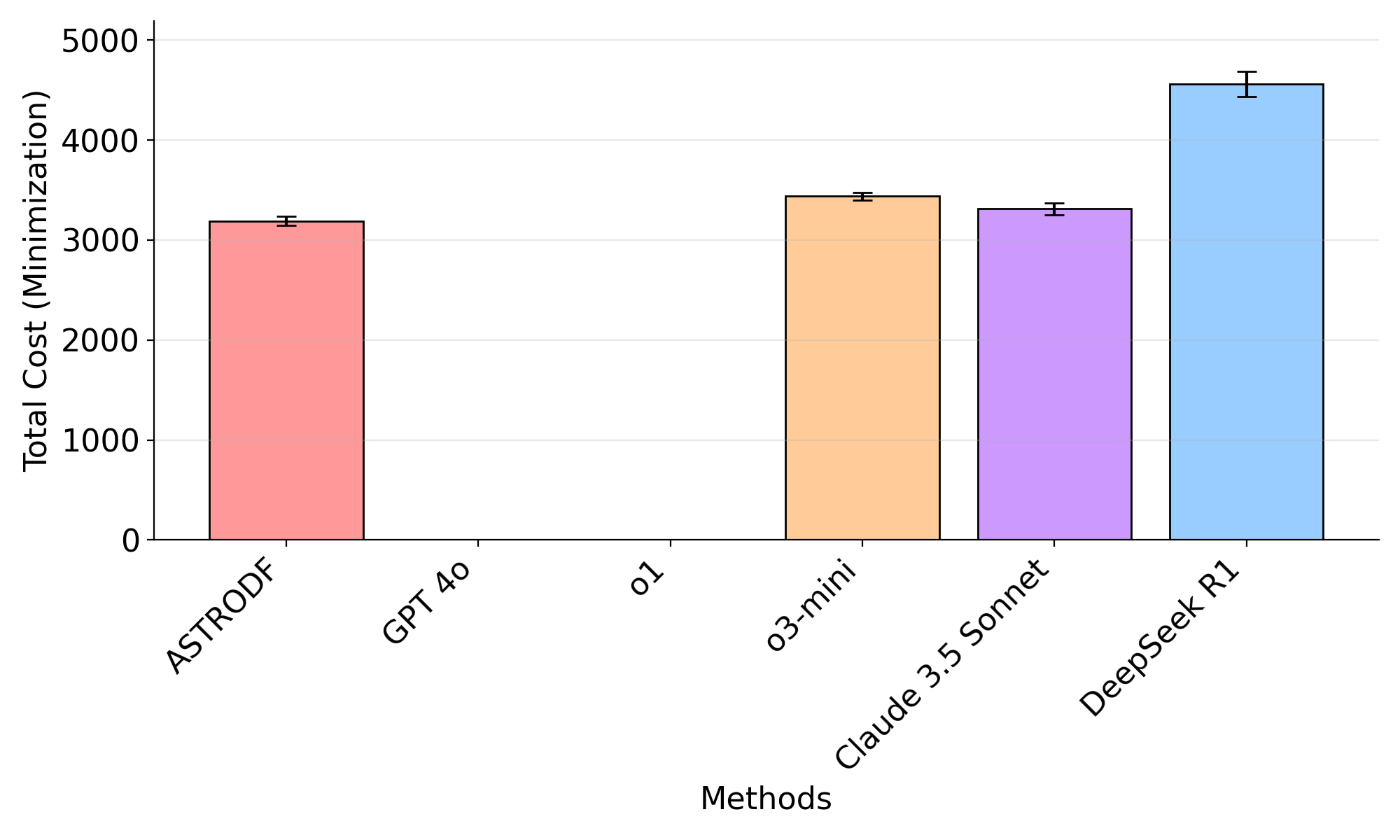}
        \caption{Dual Sourcing}
    \end{subfigure}
    \begin{subfigure}{0.48\textwidth}
    \includegraphics[width = \linewidth]{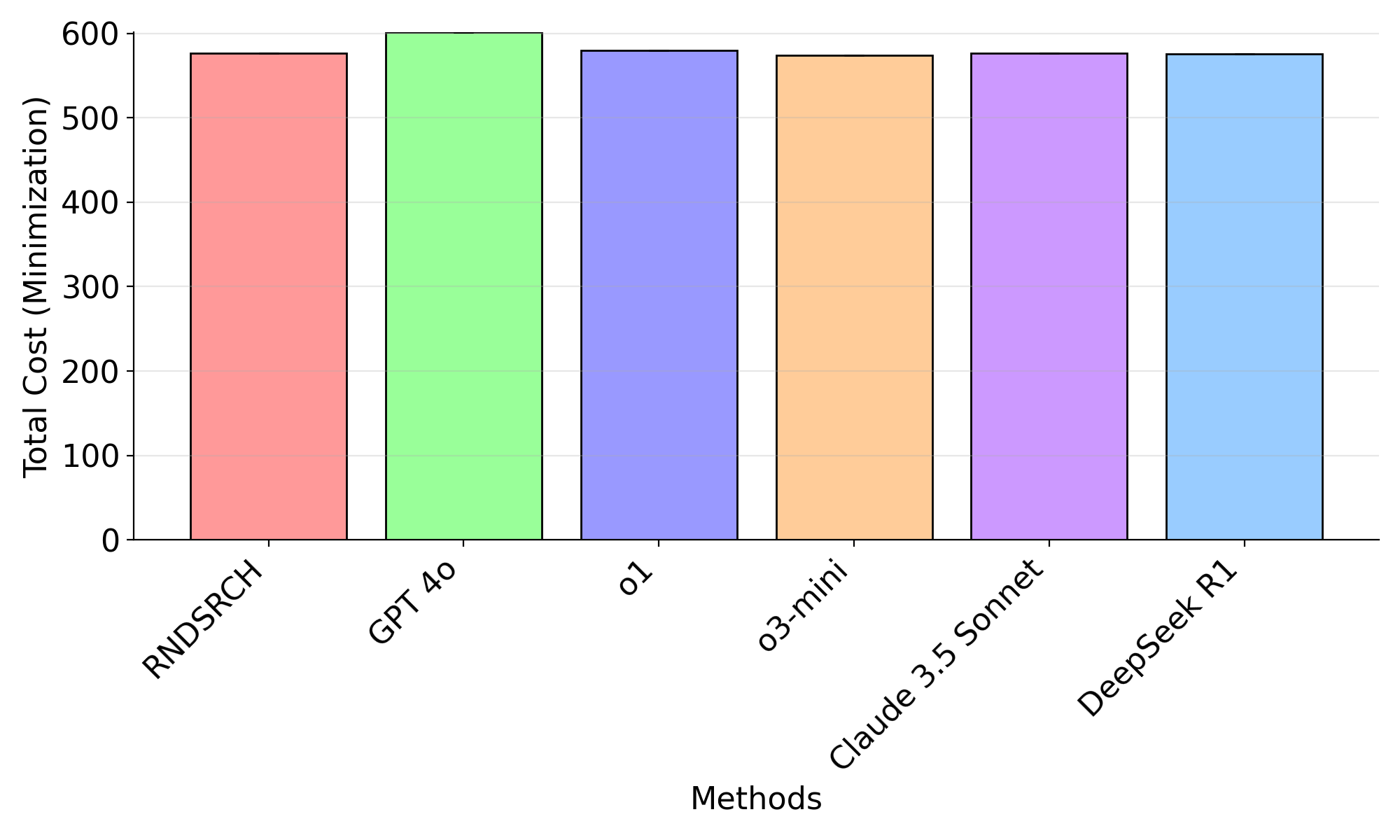}
        \caption{Facility Sizing}
    \end{subfigure}
    \caption{Performance of different LLMs for simulation-optimization problems. We omit IronOre due to differences in how LLMs implement the simulation environment, making the results incomparable. We also exclude ParamEsti, as all LLMs converge to the optimal solution.}
    \label{fig:simopt-problem-performances}
\end{figure}

In Figure \ref{fig:simopt-problem-performances}, we showcase the performance of the solution proposed by the different models. If the performance of a model is missing, it means that the model failed to produce reasonable code despite multiple attempts. Below, we summarize our main findings, followed by problem-specific analyses.
\begin{enumerate}
    \item Claude 3.5 Sonnet delivers the strongest overall performance, achieving near-optimal solutions for five out of six problems. On the other hand, despite their theoretical exam prowess, GPT-4o critically fails on the textbook Continuous Newsvendor problem and o1 can even fail to generate numerical solutions.
    \item Consistent methodological preferences surface: Claude leverages binary/differential evolution, DeepSeek-R1 defaults to coordinate descent, and o1 employs domain-constrained grid searches.
    \item The IronOre problem highlights implementation limitations, with all models producing incomparable solutions.
\end{enumerate}

As evidenced by comparing LLMs' performance on paper-based exams and on these practical problems, the top performer in one scenario may not be the best in the other. State-of-the-art models like GPT-4o or o1 may even fail to reliably produce numerical solutions. The IronOre problem also suggests that the comparison between models still needs human supervision. Overall, these observations suggest that more work is needed to reliably automate the stochastic modeling pipeline.

\subsection*{Chess Matchmaking (\href{https://simopt.readthedocs.io/en/latest/chessmm.html}{ChessMM})}\label{sec:chessmm}

This problem involves matching players on an online chess platform to minimize the average Elo difference between matched pairs, while ensuring that the average waiting time does not exceed a specified threshold ($\delta=5.0$). In this setting, players arrive according to a Poisson process and their Elo ratings are sampled from a truncated normal distribution over the interval $[0,2400]$. No theoretical optimal solution is known, and {\tt SimOpt}'s best-performing algorithm achieved an average Elo difference of 45.1246. All evaluations are conducted with a fixed budget of 1000 function calls. The different LLM approaches primarily differ in their search strategies for determining the optimal allowable Elo difference threshold, $x$. For example, GPT-4o's solution implements a line search over the full range $[0,2400]$. In contrast, o1's approach confines its grid search to the interval $[0,300]$, effectively focusing its limited budget on a more promising subset of the parameter space, which results in better performance relative to the other methods. Similarly, o3-mini's approach uses grid search but over a wider interval $[10,2400]$, which can lead to a less concentrated search and suboptimal tuning. Meanwhile, Claude's solution employs a binary search over $[0,2400]$, efficiently honing in on the smallest threshold that meets the waiting time constraint, while DeepSeek also utilizes grid search over the full range. These differences in search range and methodology underscore how a more targeted exploration---such as the one adopted by o1---can yield superior performance when the computational budget is fixed.

\subsection*{Continuous Newsvendor (\href{https://simopt.readthedocs.io/en/latest/cntnv.html}{CntNv})}
\label{sec:cntnv}
This problem considers a vendor who orders a fixed quantity of liquid at the beginning of the day. The liquid is sold to customers at a per-unit price and any unsold inventory is salvaged at a lower per-unit price, while the vendor incurs a per-unit ordering cost. In this classic formulation the optimal order quantity is known in closed form. In our evaluation, we compare ASTRO-DF alongside LLM-derived solutions, all using a fixed budget of 1000 function evaluations. Among the LLM solutions, GPT-4o failed to produce a reasonable solution, whereas o1 and o3-mini both leveraged the closed-form solution to guide their search. Specifically, o1 combined the closed-form reasoning with a random search strategy, while o3-mini employed a grid search over candidate order quantities. In contrast, Claude utilized a Nelder-Mead algorithm to iteratively converge to the optimum, and DeepSeek applied a grid search approach. The key difference in performance across these methods can be largely attributed to the search strategy and the range over which the optimization is performed.

\subsection*{Dual Sourcing (\href{https://simopt.readthedocs.io/en/latest/dualsourcing.html}{DualSourcing})}
This problem requires choosing optimal order-up-to levels for two procurement channels---one regular (lower cost, longer lead time) and one expedited (higher cost, shorter lead time)---so as to minimize the total expected daily cost (comprising holding, penalty, and ordering costs) over a simulation horizon of \(n\) periods. All the LLMs implemented a grid search style technique. Overall, the differences in performance across these models can be traced to how they allocate their fixed evaluation budget. o3-mini, Claude, and DeepSeek restrict the search to a plausible, domain-informed range and appropriately replicate the simulations. In contrast, GPT-4o and o1 misinterpreted and incorrectly implemented key aspects of the inventory dynamics---the correct handling of order arrivals, backorders, and the inventory pipeline---leading them to absurd cost estimates.

\subsection*{Facility Sizing (\href{https://simopt.readthedocs.io/en/latest/facilitysizing.html}{FacSize})}
This problem requires selecting nonnegative capacities \(x_i\) for three facilities to minimize the total installation cost while ensuring that the probability of a stockout (i.e., at least one facility experiencing demand exceeding its capacity) remains below a certain level. GPT-4o uses a continuous optimization approach with \texttt{scipy.optimize.minimize}, simulating demand samples and incorporating the stockout probability as an inequality constraint; however, its estimation of the constraint is imprecise, leading to suboptimal solutions. In contrast, o1 employs a grid search over a discrete range of candidate capacities and reuses pre-generated demand samples to estimate the stockout probability for each candidate, ultimately selecting the lowest-cost solution that meets the risk constraint. o3-mini takes a different approach by parameterizing the capacities as \(x = \mu + z\,\sigma\) and performing a binary search on the scalar parameter \(z\) to efficiently identify the smallest safety margin that satisfies the stockout constraint. Claude also relies on an iterative binary search, starting with bounds based on the mean and standard deviations of the demand distribution, and refines these bounds through repeated simulations to robustly estimate the stockout probability, thereby finding a cost-effective solution. DeepSeek initially derives a solution by setting baseline capacities based on normal quantiles and then applies coordinate descent to iteratively lower each capacity while ensuring that the overall stockout probability remains within acceptable limits.

\subsection*{Iron Ore Production with Exogenous Stochastic Price (\href{https://simopt.readthedocs.io/en/latest/ironore.html}{IronOre})}
 This problem models a mine producing and selling an item (such as iron ore) on a spot market where the daily price \(P_t\) follows a truncated mean-reverting random walk. Every day the decision maker observes \(P_t\) and the current inventory level, then makes production and sales decisions according to four thresholds: \(x_1\) is the price above which production starts or continues, \(x_2\) is the inventory level above which production is halted, \(x_3\) is the price below which production is stopped, and \(x_4\) is the price above which the entire inventory is sold. Production is limited to a maximum daily amount and capacity constraints, while holding costs apply to unsold inventory. Among the SimOpt algorithms, ASTRO-DF performed the best in terms of maximizing profit. All the LLMs differ in the way they implement the simulation environment for this problem, so their numerical results are not directly comparable. Therefore, we describe only qualitatively the strategies each one uses: GPT-4o performs a random search over the continuous threshold space using \texttt{scipy.optimize.minimize}; o1 uses a grid search over a broad range of threshold candidates, applying pre-generated price paths to evaluate each policy, but its ordering of decisions appears inconsistent; o3-mini parameterizes the thresholds relative to the mean and standard deviations of the price process and uses a binary search on a safety margin parameter combined with multiple replications to refine the candidate solution; Claude employs differential evolution, running multiple simulation trials per evaluation to iteratively improve the threshold values; and DeepSeek initiates its search with a Bonferroni-based heuristic and then applies coordinate descent to iteratively adjust the thresholds.

\subsection*{Parameter Estimation (\href{https://simopt.readthedocs.io/en/latest/paramesti.html}{ParamEsti})}
This is a classical textbook problem in which the objective is to recover the unknown parameter vector \(x^* = (x_1^*, x_2^*)\) of a two-dimensional gamma distribution from i.i.d.\ observations, with the density defined over \([0,\infty) \times [0,\infty)\). Both traditional SimOpt algorithms as well as those implemented by the various LLMs converge to the optimal solution. GPT-4o adopts a standard numerical optimization approach by constructing the negative log-likelihood function and minimizing it with \texttt{scipy.optimize.minimize} (typically using an L-BFGS-B algorithm) starting from an initial guess; this method directly maximizes the likelihood based on the observed data. In contrast, o1 takes a partial closed-form approach by solving the MLE conditions using root-finding techniques such as bisection to invert the digamma function for one parameter while using a moment condition for the other, thereby leveraging analytical properties of the gamma distribution. o3-mini employs a direct search strategy by randomly generating candidate solutions over a prescribed domain and evaluating their average log-likelihoods over multiple replications, ultimately selecting the candidate that achieves the highest value. Claude uses a numerical optimization approach---employing methods such as differential evolution or L-BFGS-B---and further refines its estimates by bootstrapping to obtain confidence intervals. Finally, DeepSeek combines the use of analytical gradients, derived from the log-likelihood function, with gradient-based optimization (again using L-BFGS-B) to efficiently and robustly recover the parameters. Despite the different methodologies, all of these approaches converge to the true parameter values as is expected in this classical problem.

\section{CONCLUSIONS} \label{sec:disc}
In this work, we evaluated LLMs' abilities to solve stochastic modeling problems through a series of homework problems, qualification-exam problems, and simulation-optimization problems. Our findings suggest that these models have the potential to automate the stochastic modeling pipeline at a level comparable with human experts, and we hope this work will inspire future research in developing intelligent OR agents that can help make real-world decisions reliably and at scale.

{
\bibliographystyle{ims}
\bibliography{bib}

\begin{thebibliography}{36}
\expandafter\ifx\csname natexlab\endcsname\relax\def\natexlab#1{#1}\fi
\expandafter\ifx\csname url\endcsname\relax
  \def\url#1{\texttt{#1}}\fi
\expandafter\ifx\csname urlprefix\endcsname\relax\def\urlprefix{URL }\fi

\bibitem[{AhmadiTeshnizi et~al.(2024)AhmadiTeshnizi, Gao, Brunborg, Talaei and Udell}]{ahmaditeshniziOptiMUS03UsingLarge2024}
\textsc{AhmadiTeshnizi, A.}, \textsc{Gao, W.}, \textsc{Brunborg, H.}, \textsc{Talaei, S.} and \textsc{Udell, M.} (2024).
\newblock {{OptiMUS-0}}.3: {{Using Large Language Models}} to {{Model}} and {{Solve Optimization Problems}} at {{Scale}}.

\bibitem[{Astorga et~al.(2024)Astorga, Liu, Xiao and van~der Schaar}]{astorgaAutoformulationMathematicalOptimization2024}
\textsc{Astorga, N.}, \textsc{Liu, T.}, \textsc{Xiao, Y.} and \textsc{van~der Schaar, M.} (2024).
\newblock Autoformulation of {{Mathematical Optimization Models Using LLMs}}.

\bibitem[{Bansal et~al.(2019)Bansal, Loos, Rabe, Szegedy and Wilcox}]{bansalHOListEnvironmentMachine2019}
\textsc{Bansal, K.}, \textsc{Loos, S.~M.}, \textsc{Rabe, M.~N.}, \textsc{Szegedy, C.} and \textsc{Wilcox, S.} (2019).
\newblock {{HOList}}: {{An Environment}} for {{Machine Learning}} of {{Higher-Order Theorem Proving}}.

\bibitem[{Bertsimas et~al.(2011)Bertsimas, Brown and Caramanis}]{bertsimas2011theory}
\textsc{Bertsimas, D.}, \textsc{Brown, D.~B.} and \textsc{Caramanis, C.} (2011).
\newblock Theory and applications of robust optimization.
\newblock \textit{SIAM review} \textbf{53} 464--501.

\bibitem[{Brown et~al.(2020)Brown, Mann, Ryder, Subbiah, Kaplan, Dhariwal, Neelakantan, Shyam, Sastry, Askell, Agarwal, {Herbert-Voss}, Krueger, Henighan, Child, Ramesh, Ziegler, Wu, Winter, Hesse, Chen, Sigler, Litwin, Gray, Chess, Clark, Berner, McCandlish, Radford, Sutskever and Amodei}]{brownLanguageModelsAre2020}
\textsc{Brown, T.~B.}, \textsc{Mann, B.}, \textsc{Ryder, N.}, \textsc{Subbiah, M.}, \textsc{Kaplan, J.}, \textsc{Dhariwal, P.}, \textsc{Neelakantan, A.}, \textsc{Shyam, P.}, \textsc{Sastry, G.}, \textsc{Askell, A.}, \textsc{Agarwal, S.}, \textsc{{Herbert-Voss}, A.}, \textsc{Krueger, G.}, \textsc{Henighan, T.}, \textsc{Child, R.}, \textsc{Ramesh, A.}, \textsc{Ziegler, D.~M.}, \textsc{Wu, J.}, \textsc{Winter, C.}, \textsc{Hesse, C.}, \textsc{Chen, M.}, \textsc{Sigler, E.}, \textsc{Litwin, M.}, \textsc{Gray, S.}, \textsc{Chess, B.}, \textsc{Clark, J.}, \textsc{Berner, C.}, \textsc{McCandlish, S.}, \textsc{Radford, A.}, \textsc{Sutskever, I.} and \textsc{Amodei, D.} (2020).
\newblock Language {{Models}} are {{Few-Shot Learners}}.

\bibitem[{Chen et~al.(2021)Chen, Tworek, Jun, Yuan, Pinto, Kaplan, Edwards, Burda, Joseph, Brockman, Ray, Puri, Krueger, Petrov, Khlaaf, Sastry, Mishkin, Chan, Gray, Ryder, Pavlov, Power, Kaiser, Bavarian, Winter, Tillet, Such, Cummings, Plappert, Chantzis, Barnes, {Herbert-Voss}, Guss, Nichol, Paino, Tezak, Tang, Babuschkin, Balaji, Jain, Saunders, Hesse, Carr, Leike, Achiam, Misra, Morikawa, Radford, Knight, Brundage, Murati, Mayer, Welinder, McGrew, Amodei, McCandlish, Sutskever and Zaremba}]{chenEvaluatingLargeLanguage2021}
\textsc{Chen, M.}, \textsc{Tworek, J.}, \textsc{Jun, H.}, \textsc{Yuan, Q.}, \textsc{Pinto, H. P. d.~O.}, \textsc{Kaplan, J.}, \textsc{Edwards, H.}, \textsc{Burda, Y.}, \textsc{Joseph, N.}, \textsc{Brockman, G.}, \textsc{Ray, A.}, \textsc{Puri, R.}, \textsc{Krueger, G.}, \textsc{Petrov, M.}, \textsc{Khlaaf, H.}, \textsc{Sastry, G.}, \textsc{Mishkin, P.}, \textsc{Chan, B.}, \textsc{Gray, S.}, \textsc{Ryder, N.}, \textsc{Pavlov, M.}, \textsc{Power, A.}, \textsc{Kaiser, L.}, \textsc{Bavarian, M.}, \textsc{Winter, C.}, \textsc{Tillet, P.}, \textsc{Such, F.~P.}, \textsc{Cummings, D.}, \textsc{Plappert, M.}, \textsc{Chantzis, F.}, \textsc{Barnes, E.}, \textsc{{Herbert-Voss}, A.}, \textsc{Guss, W.~H.}, \textsc{Nichol, A.}, \textsc{Paino, A.}, \textsc{Tezak, N.}, \textsc{Tang, J.}, \textsc{Babuschkin, I.}, \textsc{Balaji, S.}, \textsc{Jain, S.}, \textsc{Saunders, W.}, \textsc{Hesse, C.}, \textsc{Carr, A.~N.}, \textsc{Leike, J.}, \textsc{Achiam, J.}, \textsc{Misra, V.}, \textsc{Morikawa, E.}, \textsc{Radford, A.},
  \textsc{Knight, M.}, \textsc{Brundage, M.}, \textsc{Murati, M.}, \textsc{Mayer, K.}, \textsc{Welinder, P.}, \textsc{McGrew, B.}, \textsc{Amodei, D.}, \textsc{McCandlish, S.}, \textsc{Sutskever, I.} and \textsc{Zaremba, W.} (2021).
\newblock Evaluating {{Large Language Models Trained}} on {{Code}}.

\bibitem[{Cobbe et~al.(2021)Cobbe, Kosaraju, Bavarian, Chen, Jun, Kaiser, Plappert, Tworek, Hilton, Nakano, Hesse and Schulman}]{cobbeTrainingVerifiersSolve2021}
\textsc{Cobbe, K.}, \textsc{Kosaraju, V.}, \textsc{Bavarian, M.}, \textsc{Chen, M.}, \textsc{Jun, H.}, \textsc{Kaiser, L.}, \textsc{Plappert, M.}, \textsc{Tworek, J.}, \textsc{Hilton, J.}, \textsc{Nakano, R.}, \textsc{Hesse, C.} and \textsc{Schulman, J.} (2021).
\newblock Training {{Verifiers}} to {{Solve Math Word Problems}}.

\bibitem[{Dai and Harrison(2020)}]{daiProcessingNetworksFluid2020}
\textsc{Dai, J.~G.} and \textsc{Harrison, J.~M.} (2020).
\newblock \textit{Processing {{Networks}}: {{Fluid Models}} and {{Stability}}}.
\newblock 1st ed. Cambridge University Press.

\bibitem[{Dong et~al.(2017)Dong, Eckman, Zhao, Henderson and Poloczek}]{dongEmpiricallyComparingFinitetime2017}
\textsc{Dong, N.~A.}, \textsc{Eckman, D.~J.}, \textsc{Zhao, X.}, \textsc{Henderson, S.~G.} and \textsc{Poloczek, M.} (2017).
\newblock Empirically comparing the finite-time performance of simulation-optimization algorithms.
\newblock In \textit{2017 {{Winter Simulation Conference}} ({{WSC}})}. IEEE, Las Vegas, NV.

\bibitem[{Eckman et~al.(2024)Eckman, Henderson, Shashaani and Pasupathy}]{simoptgithub}
\textsc{Eckman, D.~J.}, \textsc{Henderson, S.~G.}, \textsc{Shashaani, S.} and \textsc{Pasupathy, R.} (2024).
\newblock {SimOpt}.
\newblock \url{https://github.com/simopt-admin/simopt}.

\bibitem[{Gao et~al.(2023)Gao, Madaan, Zhou, Alon, Liu, Yang, Callan and Neubig}]{gaoPALProgramaidedLanguage2023}
\textsc{Gao, L.}, \textsc{Madaan, A.}, \textsc{Zhou, S.}, \textsc{Alon, U.}, \textsc{Liu, P.}, \textsc{Yang, Y.}, \textsc{Callan, J.} and \textsc{Neubig, G.} (2023).
\newblock {{PAL}}: {{Program-aided Language Models}}.

\bibitem[{Glazer et~al.(2024)Glazer, Erdil, Besiroglu, Chicharro, Chen, Gunning, Olsson, Denain, Ho, Santos, J{\"a}rviniemi, Barnett, Sandler, Vrzala, Sevilla, Ren, Pratt, Levine, Barkley, Stewart, Grechuk, Grechuk, Enugandla and Wildon}]{glazerFrontierMathBenchmarkEvaluating2024}
\textsc{Glazer, E.}, \textsc{Erdil, E.}, \textsc{Besiroglu, T.}, \textsc{Chicharro, D.}, \textsc{Chen, E.}, \textsc{Gunning, A.}, \textsc{Olsson, C.~F.}, \textsc{Denain, J.-S.}, \textsc{Ho, A.}, \textsc{Santos, E. d.~O.}, \textsc{J{\"a}rviniemi, O.}, \textsc{Barnett, M.}, \textsc{Sandler, R.}, \textsc{Vrzala, M.}, \textsc{Sevilla, J.}, \textsc{Ren, Q.}, \textsc{Pratt, E.}, \textsc{Levine, L.}, \textsc{Barkley, G.}, \textsc{Stewart, N.}, \textsc{Grechuk, B.}, \textsc{Grechuk, T.}, \textsc{Enugandla, S.~V.} and \textsc{Wildon, M.} (2024).
\newblock {{FrontierMath}}: {{A Benchmark}} for {{Evaluating Advanced Mathematical Reasoning}} in {{AI}}.

\bibitem[{Hendrycks et~al.(2021)Hendrycks, Burns, Kadavath, Arora, Basart, Tang, Song and Steinhardt}]{hendrycksMeasuringMathematicalProblem2021}
\textsc{Hendrycks, D.}, \textsc{Burns, C.}, \textsc{Kadavath, S.}, \textsc{Arora, A.}, \textsc{Basart, S.}, \textsc{Tang, E.}, \textsc{Song, D.} and \textsc{Steinhardt, J.} (2021).
\newblock Measuring {{Mathematical Problem Solving With}} the {{MATH Dataset}}.

\bibitem[{Huang et~al.(2025)Huang, Tang, Hu, Jiang, Zheng, Ge, Wang and Wang}]{huangORLMCustomizableFramework2025}
\textsc{Huang, C.}, \textsc{Tang, Z.}, \textsc{Hu, S.}, \textsc{Jiang, R.}, \textsc{Zheng, X.}, \textsc{Ge, D.}, \textsc{Wang, B.} and \textsc{Wang, Z.} (2025).
\newblock {{ORLM}}: {{A Customizable Framework}} in {{Training Large Models}} for {{Automated Optimization Modeling}}.

\bibitem[{{INFORMS}(2025)}]{informs_edelman}
\textsc{{INFORMS}} (2025).
\newblock Franz edelman award for achievement in advanced analytics, operations research, and management science.
\newblock \url{https://www.informs.org/Recognizing-Excellence/INFORMS-Prizes/Franz-Edelman-Award}.
\newblock Accessed 20 June 2025.

\bibitem[{Jiang et~al.(2022)Jiang, Li, Tworkowski, Czechowski, Odrzyg{\'o}{\'z}d{\'z}, Mi{\l}o{\'s}, Wu and Jamnik}]{jiangThorWieldingHammers2022}
\textsc{Jiang, A.~Q.}, \textsc{Li, W.}, \textsc{Tworkowski, S.}, \textsc{Czechowski, K.}, \textsc{Odrzyg{\'o}{\'z}d{\'z}, T.}, \textsc{Mi{\l}o{\'s}, P.}, \textsc{Wu, Y.} and \textsc{Jamnik, M.} (2022).
\newblock Thor: {{Wielding Hammers}} to {{Integrate Language Models}} and {{Automated Theorem Provers}}.

\bibitem[{Jiang et~al.(2025)Jiang, Shu, Qian, Lu, Zhou, Zhou and Yu}]{jiangLLMOPTLearningDefine2025}
\textsc{Jiang, C.}, \textsc{Shu, X.}, \textsc{Qian, H.}, \textsc{Lu, X.}, \textsc{Zhou, J.}, \textsc{Zhou, A.} and \textsc{Yu, Y.} (2025).
\newblock {{LLMOPT}}: {{Learning}} to {{Define}} and {{Solve General Optimization Problems}} from {{Scratch}}.

\bibitem[{Kaliszyk et~al.(2017)Kaliszyk, Chollet and Szegedy}]{kaliszykHolStepMachineLearning2017}
\textsc{Kaliszyk, C.}, \textsc{Chollet, F.} and \textsc{Szegedy, C.} (2017).
\newblock {{HolStep}}: {{A Machine Learning Dataset}} for {{Higher-order Logic Theorem Proving}}.

\bibitem[{Li et~al.(2023)Li, Zhang and {Mak-Hau}}]{liSynthesizingMixedintegerLinear2023}
\textsc{Li, Q.}, \textsc{Zhang, L.} and \textsc{{Mak-Hau}, V.} (2023).
\newblock Synthesizing mixed-integer linear programming models from natural language descriptions.

\bibitem[{Li et~al.(2022)Li, Choi, Chung, Kushman, Schrittwieser, Leblond, Eccles, Keeling, Gimeno, Lago, Hubert, Choy, {d'Autume}, Babuschkin, Chen, Huang, Welbl, Gowal, Cherepanov, Molloy, Mankowitz, Robson, Kohli, de~Freitas, Kavukcuoglu and Vinyals}]{liCompetitionLevelCodeGeneration2022}
\textsc{Li, Y.}, \textsc{Choi, D.}, \textsc{Chung, J.}, \textsc{Kushman, N.}, \textsc{Schrittwieser, J.}, \textsc{Leblond, R.}, \textsc{Eccles, T.}, \textsc{Keeling, J.}, \textsc{Gimeno, F.}, \textsc{Lago, A.~D.}, \textsc{Hubert, T.}, \textsc{Choy, P.}, \textsc{{d'Autume}, C. d.~M.}, \textsc{Babuschkin, I.}, \textsc{Chen, X.}, \textsc{Huang, P.-S.}, \textsc{Welbl, J.}, \textsc{Gowal, S.}, \textsc{Cherepanov, A.}, \textsc{Molloy, J.}, \textsc{Mankowitz, D.~J.}, \textsc{Robson, E.~S.}, \textsc{Kohli, P.}, \textsc{de~Freitas, N.}, \textsc{Kavukcuoglu, K.} and \textsc{Vinyals, O.} (2022).
\newblock Competition-{{Level Code Generation}} with {{AlphaCode}}.
\newblock \textit{Science} \textbf{378} 1092--1097.

\bibitem[{Mostajabdaveh et~al.(2024)Mostajabdaveh, Yu, Ramamonjison, Carenini, Zhou and Zhang}]{mostajabdavehOptimizationModelingVerification2024}
\textsc{Mostajabdaveh, M.}, \textsc{Yu, T.~T.}, \textsc{Ramamonjison, R.}, \textsc{Carenini, G.}, \textsc{Zhou, Z.} and \textsc{Zhang, Y.} (2024).
\newblock Optimization modeling and verification from problem specifications using a multi-agent multi-stage {{LLM}} framework.
\newblock \textit{INFOR: Information Systems and Operational Research} \textbf{62} 599--617.

\bibitem[{Polu and Sutskever(2020)}]{poluGenerativeLanguageModeling2020}
\textsc{Polu, S.} and \textsc{Sutskever, I.} (2020).
\newblock Generative {{Language Modeling}} for {{Automated Theorem Proving}}.

\bibitem[{Ramamonjison et~al.(2022)Ramamonjison, Li, Yu, He, Rengan, {Banitalebi-Dehkordi}, Zhou and Zhang}]{ramamonjisonAugmentingOperationsResearch2022}
\textsc{Ramamonjison, R.}, \textsc{Li, H.}, \textsc{Yu, T.~T.}, \textsc{He, S.}, \textsc{Rengan, V.}, \textsc{{Banitalebi-Dehkordi}, A.}, \textsc{Zhou, Z.} and \textsc{Zhang, Y.} (2022).
\newblock Augmenting {{Operations Research}} with {{Auto-Formulation}} of {{Optimization Models}} from {{Problem Descriptions}}.

\bibitem[{Rockt{\"a}schel and Riedel(2017)}]{rocktaschelEndtoEndDifferentiableProving2017}
\textsc{Rockt{\"a}schel, T.} and \textsc{Riedel, S.} (2017).
\newblock End-to-{{End Differentiable Proving}}.

\bibitem[{{Romera-Paredes} et~al.(2024){Romera-Paredes}, Barekatain, Novikov, Balog, Kumar, Dupont, Ruiz, Ellenberg, Wang, Fawzi, Kohli and Fawzi}]{romera-paredesMathematicalDiscoveriesProgram2024}
\textsc{{Romera-Paredes}, B.}, \textsc{Barekatain, M.}, \textsc{Novikov, A.}, \textsc{Balog, M.}, \textsc{Kumar, M.~P.}, \textsc{Dupont, E.}, \textsc{Ruiz, F. J.~R.}, \textsc{Ellenberg, J.~S.}, \textsc{Wang, P.}, \textsc{Fawzi, O.}, \textsc{Kohli, P.} and \textsc{Fawzi, A.} (2024).
\newblock Mathematical discoveries from program search with large language models.
\newblock \textit{Nature} \textbf{625} 468--475.

\bibitem[{Schick et~al.(2023)Schick, {Dwivedi-Yu}, Dess{\`i}, Raileanu, Lomeli, Zettlemoyer, Cancedda and Scialom}]{schickToolformerLanguageModels2023}
\textsc{Schick, T.}, \textsc{{Dwivedi-Yu}, J.}, \textsc{Dess{\`i}, R.}, \textsc{Raileanu, R.}, \textsc{Lomeli, M.}, \textsc{Zettlemoyer, L.}, \textsc{Cancedda, N.} and \textsc{Scialom, T.} (2023).
\newblock Toolformer: {{Language Models Can Teach Themselves}} to {{Use Tools}}.

\bibitem[{Shinn et~al.(2023)Shinn, Cassano, Gopinath, Narasimhan and Yao}]{shinn2023reflexion}
\textsc{Shinn, N.}, \textsc{Cassano, F.}, \textsc{Gopinath, A.}, \textsc{Narasimhan, K.} and \textsc{Yao, S.} (2023).
\newblock Reflexion: Language agents with verbal reinforcement learning.
\newblock \textit{Advances in Neural Information Processing Systems} \textbf{36} 8634--8652.

\bibitem[{Wang et~al.(2023)Wang, Xie, Jiang, Mandlekar, Xiao, Zhu, Fan and Anandkumar}]{wang2023voyager}
\textsc{Wang, G.}, \textsc{Xie, Y.}, \textsc{Jiang, Y.}, \textsc{Mandlekar, A.}, \textsc{Xiao, C.}, \textsc{Zhu, Y.}, \textsc{Fan, L.} and \textsc{Anandkumar, A.} (2023).
\newblock Voyager: An open-ended embodied agent with large language models.
\newblock \textit{arXiv preprint arXiv:2305.16291} .

\bibitem[{Wei et~al.(2023)Wei, Wang, Schuurmans, Bosma, Ichter, Xia, Chi, Le and Zhou}]{weiChainofThoughtPromptingElicits2023}
\textsc{Wei, J.}, \textsc{Wang, X.}, \textsc{Schuurmans, D.}, \textsc{Bosma, M.}, \textsc{Ichter, B.}, \textsc{Xia, F.}, \textsc{Chi, E.}, \textsc{Le, Q.} and \textsc{Zhou, D.} (2023).
\newblock Chain-of-{{Thought Prompting Elicits Reasoning}} in {{Large Language Models}}.

\bibitem[{Xiao et~al.(2024)Xiao, Zhang, Wu, Xu, Wang, Han, Fu, Zhong, Zeng, Song and Chen}]{xiaoCHAINOFEXPERTSWHENLLMS2024}
\textsc{Xiao, Z.}, \textsc{Zhang, D.}, \textsc{Wu, Y.}, \textsc{Xu, L.}, \textsc{Wang, Y.}, \textsc{Han, X.}, \textsc{Fu, X.}, \textsc{Zhong, T.}, \textsc{Zeng, J.}, \textsc{Song, M.} and \textsc{Chen, G.} (2024).
\newblock {{Chain-of-Experts}}: {{When LLMs Meet Complex Operations Research Problems}}.
\newblock \textit{ICLR} .

\bibitem[{Yang and Deng(2019)}]{yangLearningProveTheorems2019}
\textsc{Yang, K.} and \textsc{Deng, J.} (2019).
\newblock Learning to {{Prove Theorems}} via {{Interacting}} with {{Proof Assistants}}.

\bibitem[{Yang et~al.(2023)Yang, Swope, Gu, Chalamala, Song, Yu, Godil, Prenger and Anandkumar}]{yangLeanDojoTheoremProving2023}
\textsc{Yang, K.}, \textsc{Swope, A.~M.}, \textsc{Gu, A.}, \textsc{Chalamala, R.}, \textsc{Song, P.}, \textsc{Yu, S.}, \textsc{Godil, S.}, \textsc{Prenger, R.} and \textsc{Anandkumar, A.} (2023).
\newblock {{LeanDojo}}: {{Theorem Proving}} with {{Retrieval-Augmented Language Models}}.

\bibitem[{Yao et~al.(2023)Yao, Zhao, Yu, Du, Shafran, Narasimhan and Cao}]{yao2023react}
\textsc{Yao, S.}, \textsc{Zhao, J.}, \textsc{Yu, D.}, \textsc{Du, N.}, \textsc{Shafran, I.}, \textsc{Narasimhan, K.} and \textsc{Cao, Y.} (2023).
\newblock React: Synergizing reasoning and acting in language models.
\newblock In \textit{International Conference on Learning Representations (ICLR)}.

\bibitem[{Zhang and Luo(2025)}]{zhang2025or}
\textsc{Zhang, B.} and \textsc{Luo, P.} (2025).
\newblock Or-llm-agent: Automating modeling and solving of operations research optimization problem with reasoning large language model.
\newblock \textit{arXiv preprint arXiv:2503.10009} .

\bibitem[{Zheng et~al.(2022)Zheng, Han and Polu}]{zhengMiniF2FCrosssystemBenchmark2022}
\textsc{Zheng, K.}, \textsc{Han, J.~M.} and \textsc{Polu, S.} (2022).
\newblock {{MiniF2F}}: A cross-system benchmark for formal {{Olympiad-level}} mathematics.

\bibitem[{Zheng et~al.(2023)Zheng, Chiang, Sheng, Zhuang, Wu, Zhuang, Lin, Li, Li, Xing, Zhang, Gonzalez and Stoica}]{zhengJudgingLLMasaJudgeMTBench2023}
\textsc{Zheng, L.}, \textsc{Chiang, W.-L.}, \textsc{Sheng, Y.}, \textsc{Zhuang, S.}, \textsc{Wu, Z.}, \textsc{Zhuang, Y.}, \textsc{Lin, Z.}, \textsc{Li, Z.}, \textsc{Li, D.}, \textsc{Xing, E.~P.}, \textsc{Zhang, H.}, \textsc{Gonzalez, J.~E.} and \textsc{Stoica, I.} (2023).
\newblock Judging {{LLM-as-a-Judge}} with {{MT-Bench}} and {{Chatbot Arena}}.

\end{thebibliography}
}

\newpage 
\appendix
\section{Example Homework Problems}\label{sec:example}
In this section, for the three example problems given in Section \ref{sec:hw}, we provide the reference solutions, GPT-4o's answers, and evaluations from the LLM judge. 

\subsection{Example 1: A Probability Theory Problem}
\begin{examplebox}[Problem Statement]
Consider a sequence $X_{1}, X_{2}, \ldots$, of i.i.d. random variables, each having an exponential distribution with parameter 1. Let $M_{n}:=\max \left\{X_{1}, \ldots, X_{n}\right\}$. \\
(a) Let $Y_{n}=X_{n} \mathbf{1}\left\{X_{n} \leq \log n\right\}$ denote a truncated exponential, i.e., $Y_{n}=X_{n}$ if $X_{n} \leq \log n$ and equals zero otherwise. Prove that $Y_{n} \neq X_{n}$, i.o., almost surely. \\
(b) Prove that $M_{n} / \log n \rightarrow 1$ almost surely as $n \rightarrow \infty$, where $\log$ denotes the natural logarithm.
\end{examplebox}

\begin{examplebox}[Reference Solution]
(a) Consider
$$
\sum_{n=1}^{\infty} \mathrm{P}\left\{X_{n} >\log n\right\}=\sum_{n=1}^{\infty}\frac{1}{n}=\infty.
$$
By Borel-Cantelli Lemma, it follows that
$$
\mathrm{P}\left\{X_{n} > \log n \text { i.o.}\right\}=1
$$
With $Y_{n} = X_{n} 1\left\{\left|X_{n}\right| \leq \log n\right\}$, it is clear that
$$
\left\{\omega: X_{n}(\omega) > \log n \text { i.o.}\right\} \subseteq\left\{\omega: Y_{n}(\omega) \ne X_{n}(\omega) \text { i.o.}\right\}.
$$
Since the event on the left-hand side has probability one, so does the event on the righthand side. That is, $Y_{n} \ne X_{n}$ infinitely often, almost surely.\\
\vspace{0.5em}

\noindent (b) Fix $\varepsilon>0$. We first show that 
$$
\left\{M_{n} \leq(1+\varepsilon) \log n \text{ ev.}\right\}=\left\{X_{n} \leq(1+\varepsilon) \log n \text{ ev.}\right\}
$$
For every $\omega \in\left\{X_{n}(\omega) \leq(1+\varepsilon) \log n \text{ ev.}\right\}$, we have
$$
\exists N(\omega) \in \NN, \quad X_{n}(\omega) \leq(1+\varepsilon) \log n \quad \text { for all } n \geq N(\omega).
$$
It follows that
$$
\max _{1 \leq i \leq n} X_{i}(\omega) \leq(1+\varepsilon) \log n \quad \text { for all } n>\max \left\{N(\omega), e^{\frac{1}{1+\varepsilon} \max _{1 \leq i \leq N(\omega)} X_{i}(\omega)}\right\}.
$$
In other words,
$$
M_{n}(\omega) \leq(1+\varepsilon) \log n, \text { eventually. }
$$
Therefore, $\left\{X_{n} \leq(1+\varepsilon) \log n \text{ ev.}\right\}\subseteq\left\{M_{n} \leq(1+\varepsilon) \log n \text{ ev.}\right\}$. It is trivial to see that 
$$
\left\{M_{n} \leq(1+\varepsilon) \log n \text{ ev.}\right\} \subseteq\left\{X_{n} \leq(1+\varepsilon) \log n \text{ ev.}\right\},
$$
so the two events are in fact equivalent. Consequently, their complements are equivalent, that is,
$$
\left\{M_{n}>(1+\varepsilon) \log n \text { i.o.}\right\}=\left\{X_{n}>(1+\varepsilon) \log n \text { i.o.}\right\}.
$$
By Borel-Cantelli Lemma,
$$
\sum_{n=1}^{\infty} \mathrm{P}\left\{X_{n} \geq(1+\varepsilon) \log n\right\}=\sum_{n=1}^{\infty} \frac{1}{n^{1+\varepsilon}}<\infty \Rightarrow \mathrm{P}\left\{X_{n} \geq(1+\varepsilon) \log n \text { i.o.}\right\}=0.
$$
Therefore,
\begin{equation}\label{eq:probtheory1}
\mathrm{P}\left\{\frac{M_{n}}{\log n}>1+\varepsilon \text { i.o.}\right\}=0, \quad \forall \varepsilon>0.
\end{equation}
Recall that
$$
\mathrm{P}\left(M_{n} \leq x\right)=\mathrm{P}\left(\max _{1 \leq i \leq n} X_{i} \leq x\right)=\prod_{i=1}^{n} \mathrm{P}\left(X_{i} \leq x\right)=\left(1-e^{-x}\right)^{n},
$$
so for large $n$,
$$
\mathrm{P}\left\{\frac{M_{n}}{\log n}<1-\varepsilon\right\}=\mathrm{P}\left\{M_{n}<(1-\varepsilon) \log n\right\}=\left(1-e^{-(1-\varepsilon) \log n}\right)^{n}=\left(1-\frac{1}{n^{1-\varepsilon}}\right)^{n} \approx e^{-n^{\varepsilon}}.
$$
Therefore,
$$
\sum_{n=1}^{\infty} \mathrm{P}\left\{\frac{M_{n}}{\log n}<1-\varepsilon\right\}<\infty.
$$
By Borel-Cantelli Lemma,
\begin{equation}\label{eq:probtheory2}
\mathrm{P}\left\{\frac{M_{n}}{\log n}<1-\varepsilon \text { i.o.}\right\}=0, \quad \forall \varepsilon>0.
\end{equation}
Combining \eqref{eq:probtheory1} and \eqref{eq:probtheory2}, we conclude that $M_{n} / \log n \rightarrow 1$ almost surely as $n \rightarrow \infty$.
\end{examplebox}

\begin{examplebox}[GPT-4o's Response]
To solve this problem rigorously, we will work through each part separately, providing detailed explanations and proofs.
\vspace{1em}

\noindent \textbf{(a) $\ Y_{n} \neq X_{n}$ i.o., almost surely.}

\underline{Definitions and Setup}:
The sequence $\{X_n\}_{n \geq 1}$ consists of i.i.d. random variables following an exponential distribution with parameter 1. The exponential distribution has a cumulative distribution function (CDF) given by $F(x) = 1 - e^{-x}$ for $x \geq 0$. The random variable $Y_n = X_n \mathbb{I}\{X_n \leq \log n\}$ indicates that $Y_n = X_n$ if $X_n \leq \log n$, and $Y_n = 0$ otherwise. We need to show that $Y_n \neq X_n$ occurs infinitely often (i.o.), almost surely.

\underline{Approach}:
For $Y_n \neq X_n$, it must be that $X_n > \log n$. We will use the Borel-Cantelli lemma to show that this event ($\{X_n > \log n\}$) happens infinitely often with probability 1.

\underline{Probability of Event}:
The probability that $X_n > \log n$ is given by:
\[ P(X_n > \log n) = 1 - P(X_n \leq \log n) = e^{-\log n} = \frac{1}{n}. \]

\underline{Series Sum}:
Consider the series:
\[ \sum_{n=1}^{\infty} P(X_n > \log n) = \sum_{n=1}^{\infty} \frac{1}{n}, \]
which is the harmonic series and diverges.

\underline{Application of Borel-Cantelli Lemma}:
Since the series diverges, by the Borel-Cantelli lemma, $P(X_n > \log n \ \text{i.o.}) = 1$. Therefore, $P(Y_n \neq X_n \ \text{i.o.}) = 1$, which means $Y_n \neq X_n$ happens infinitely often almost surely.
\vspace{1em}

\noindent \textbf{(b) $M_n / \log n \rightarrow 1$ almost surely as $n \rightarrow \infty$.}

\underline{Definitions and Setup}:
The random variable $M_n = \max\{X_1, X_2, \ldots, X_n\}$ is the maximum of the first $n$ i.i.d. exponentially distributed random variables. We aim to show that $\frac{M_n}{\log n} \to 1$ almost surely as $n \to \infty$.

\underline{Distribution of $M_n$}:
The CDF of $M_n$ is given by:
\[ F_{M_n}(x) = P(M_n \leq x) = (P(X_1 \leq x))^n = (1 - e^{-x})^n. \]

\underline{Probability for $M_n$}:
For large $n$ and letting $x = (1 + \epsilon) \log n$, the probability $M_n \leq (1 + \epsilon) \log n$ is:
\[ P(M_n \leq (1+\epsilon)\log n) = (1 - n^{-(1+\epsilon)})^n. \]
Using the approximation $(1 - \frac{1}{n^{1+\epsilon}})^n \to e^{-n^{-\epsilon}} \approx 1$ for large $n$, this probability approaches 0.

On the other hand, for $x = (1 - \epsilon) \log n$, the probability:
\[ P(M_n \leq (1-\epsilon)\log n) = (1 - n^{-(1-\epsilon)})^n \approx e^{-n^{\epsilon}}, \]
which approaches 0 for any $\epsilon > 0$.

\underline{Almost Sure Convergence}:
By this argument, $\frac{M_n}{\log n}$ does not deviate from 1 as $n$ becomes large, meaning:
\[ P\left(\frac{M_n}{\log n} \not\rightarrow 1\right) = 0. \]
Thus, by the Squeeze Theorem and convergence in probability, $\frac{M_n}{\log n} \rightarrow 1$ almost surely as $n \to \infty$.

This completes both parts of the problem, proving the required properties with detailed reasoning.
\end{examplebox}

\begin{examplebox}[LLM Judge's Evaluation]
[[80]] The student's solution is largely correct and well-reasoned, but it does contain some minor inaccuracies and lacks certain elements of rigor when compared to the model solution.

\begin{itemize}
\item \textbf{Part (a):} 
\begin{itemize}
    \item The student correctly identifies the application of the Borel-Cantelli Lemma for part (a). However, in the explanation of the setup, they include some redundant elements by trying to monotone equivalency between events, which although not incorrect, it is less straightforward than the given sketch. This part generally follows the correct steps.
    \item They correctly derive the probability expressions and apply the Borel-Cantelli lemma to show that $X_n > \log n$ occurs infinitely often. This part is well done.
\end{itemize}
\item \textbf{Part (b):}
\begin{itemize}
    \item The student accurately describes the setup by determining the distribution of $M_n$ and correctly states the CDF for $M_n$. Their computation matches the sketch solution provided.
    \item The reasoning involving limits and approximations is correct; however, when they claim ``this probability approaches 0,'' the reasoning could be clearer, noting that it should be shown that such events are complementary in the infinite sample space. The student briefly discusses the probabilities but doesn't firmly link it back to insights gained from the Borel-Cantelli Lemma. 
    \item They do not fully articulate that for convergence in almost sure sense, each of the complementary sets in sequences pertaining to $(1+\epsilon) \log n$ and $(1-\epsilon) \log n$ should be treated with an explanation similar to Borel-Cantelli (what happens with the tail distribution) for more clarity.
    \item Their explanation involving the Squeeze Theorem is slightly confusing, but the conclusion is correct. They could improve by directly relating back to almost sure convergence conditions and more clear application of inequalities.
\end{itemize}
\end{itemize}
Overall, both parts a and b are generally correct, and they exhibit a good understanding of the underlying principles. Some points are taken off due to the slight lack of detail and clarity when dealing with almost sure convergence and the rigorous tie to provided information from Borel-Cantelli.
\end{examplebox}

\begin{remark}
Overall, the judge's evaluation is on track. Part (a) indeed deserves full scores, while part (b), depending on how harsh the grader is, may receive a lower score than that given by the judge. Though part (b)'s answer is a bit different from the expected solution, the judge is correct that this approach is not wrong and the conclusion is correct. We think this is a good evaluation with a slight inflation of the score.
\end{remark}

\subsection{Example 2: A Stochastic Process Problem}
\begin{examplebox}[Problem Statement]
Consider an irreducible discrete-time Markov chain $X_n$ on a finite state space. Let $p$ denote its transition probability matrix. A function $f$ is said to be superharmonic if $f(x) \geq \sum_{y} p(x, y) f(y)$ or equivalently $f\left(X_{n}\right)$ is a supermartingale. Show that the Markov chain is recurrent if and only if every nonnegative superharmonic function is constant.
\end{examplebox}

\begin{examplebox}[Reference Solution]
Suppose that $p$ is recurrent. For any nonnegative superharmonic function $f$, we know that $f\left(X_{n}\right)$ is a nonnegative supermartingale. By martingale convergence theorem, we get that $f\left(X_{n}\right)$ has a limit. If there exist $i \neq j$ such that $f(i) \neq f(j)$, then $f\left(X_{n}\right)$ cannot converge almost surely since the chain visits both $i$ and $j$ infinitely often. Thus, $f$ must be constant. 

Now we suppose that $p$ is transient. We fix a state $x$. Then we know that $P_{x}\left(\exists n>0, X_{n}=x\right)<1$. We define $f(y)=P_{y}\left(\exists n>0, X_{n}=x\right)$ and $f(x)=1$. It is easily verified that $f$ is a nonnegative superharmonic function. However, $f$ is not constant. Otherwise, we have that $f(y)=1, \forall y \neq x$, and hence $P_{x}\left(\exists n>0, X_{n}=x\right)=1$, which is a contradiction.
\end{examplebox}

\begin{examplebox}[GPT-4o's Response]
To prove this statement, we will analyze both directions of the logical equivalence:

\noindent \textbf{1. If \( p \) is recurrent, then every nonnegative superharmonic function is constant.}

Recurrence implies that every state \( x \) is visited infinitely often with probability 1. A Markov chain with a finite state space is irreducible and recurrent if each state is positive recurrent, i.e. the expected return time to the state is finite.

Consider any nonnegative superharmonic function \( f \). By definition, a function \( f \) is superharmonic if:
\[
f(x) \geq \sum_y p(x, y) f(y)
\]
for all states \( x \). 

Since the Markov chain is recurrent, every state \( x \) is visited infinitely often. Denote by \( \mathbb{E}_x \) the expectation starting from \( x \), and consider the process \( f(X_n) \). By the optional stopping theorem applied to the bounded stopping times (since \( f \) is nonnegative and superharmonic, so the process is a supermartingale), for any state \( x \) and any stopping time \( \tau \) with finite expectation, we have:
\[
\mathbb{E}_x[f(X_{\tau})] \leq f(x).
\]
For recurrence, if we let \( \tau \) be the hitting time of any given state \( y \), then:
\[
\mathbb{E}_x[f(X_{\tau})] = f(y) \leq f(x).
\]
Since the chain is irreducible and recurrent, we can choose such \( \tau \) such that every state is recurrently revisited. Therefore, for any distinct points \( x \) and \( y \), the expected values should balance out, i.e. the only way for the inequality to hold every time is for \( f(y) = f(x) \). Since this applies for any pair of states, \( f \) must take the same constant value at every state.
\vspace{0.5em}

\noindent \textbf{2. If every nonnegative superharmonic function is constant, then \( p \) is recurrent.}

Conversely, assume every nonnegative superharmonic function is constant. We need to show this implies recurrence of \( p \).

Assume toward contradiction that \( p \) is transient. In a transient Markov chain, there exist states \( x \) such that the expected number of visits to \( x \) is finite. Define \( f(x) \) as the expected number of visits to state \( x \) starting from state \( x \). For transient states, this function \( f(x) \) is finite.

The function \( f(x) \) is indeed nonnegative superharmonic:
\[
f(x) = 1 + \sum_y p(x, y) f(y),
\]
which implies:
\[
f(x) \geq \sum_y p(x, y) f(y).
\]
The form of \( f(x) \) does not imply constancy unless all states are visited infinitely often, contradicting transience.

Thus, if every nonnegative superharmonic function is forced to be constant, then \( p \) cannot be transient and must therefore be recurrent.

This completes the proof, showing that \( p \) is recurrent if and only if every nonnegative superharmonic function is constant.
\end{examplebox}

\begin{examplebox}[LLM Judge's Evaluation]
[[85]] The student's solution demonstrates a good understanding of the problem and manages to provide a coherent argument for both directions of the proof. Here is a breakdown of the evaluation:

\begin{itemize}
\item \textbf{Correctness:}
\begin{itemize}
\item The student correctly identifies the relationship between superharmonic functions and recurrence/transience.
\item For the first direction (``If \( p \) is recurrent, then every nonnegative superharmonic function is constant''), the student effectively employs characteristics of recurrence and the optional stopping theorem to justify why \( f \) must be constant.
\item For the second direction (``If every nonnegative superharmonic function is constant, then \( p \) is recurrent''), the student constructs a correct argument using the definition of transience and proposes a function \( f \) which, under transience, would not be constant, thus obtaining a contradiction. 
\end{itemize}
\item \textbf{Completeness:}
\begin{itemize}
\item The student does a good job covering the logical implications in both directions of the proof.
\item There is subtlety in the use of the optional stopping theorem, which is treated with reasonable rigor, although some parts could be slightly clearer in terms of explaining the choice of stopping times and their finiteness.
\item A minor omission is the lack of explicit consideration of infinite state spaces as the student implicitly assumes the state space might be finite. In the context of finite state spaces, the argument holds strongly.
\end{itemize}
\item \textbf{Rigor and Clarity:}
\begin{itemize}
\item The arguments are well structured but could be improved in rigor by explicitly addressing any assumptions such as the potential boundedness of certain functions or convergence issues.
\item The explanation of why \( f \) must remain constant is solid, relying on fundamentals of supermartingale properties under the recurrence condition.
\item In the second part, while the argument against transience by showing \( f(x) = 1 + \sum_y p(x, y) f(y) \) and using superharmonic properties is effective, it would benefit from a more explicit illustration of why this function is indeed not constant under transient conditions.
\end{itemize}
\end{itemize}
Overall, the student provides a strong solution that would only benefit from slight clarifications and improvements in detailing certain steps comprehensively.
\end{examplebox}

\begin{remark}
Overall, the judge's evaluation is decent. It is clear that this answer is better than the previous one, and indeed the judge's score is higher. However, the judge made a mistake in saying that the student ``implicitly assumes the state space might be finite,'' while the problem explicitly states that the Markov chain is on a finite state space. In this case, we think the answer deserves a slightly higher score than the judge's evaluation. 
\end{remark}

\subsection{Example 3: A Stochastic Modeling Problem}
\begin{examplebox}[Problem Statement]
Consider an inventory model in which demand for a commodity arrives at the end of each day. Successive demands are i.i.d. with distribution function $F(\cdot)$. The following $(s, S)$ policy is used: if the inventory level at the beginning of a day is less than or equal to $s$, we order up to $S$, and if the level is greater than $s$ no action is taken. Orders are assumed to filled instantaneously. Let $\left\{X_{n}\right\}_{n \geq 1}$ be the inventory level at the beginning of the $n$th day, right after delivery of inventory (if any). Suppose that $X_{1}=S$.\\
(a) Prove that $X_{n}$ is regenerative with points of regeneration that are given by the order epochs.\\
(b) Let $\left\{\tau_{n}\right\}$ denote the lengths of the regenerative cycles, derive an expression for $\mathbb{E} \tau_{1}$.\\
(c) Derive an expression for $\lim _{n \rightarrow \infty} \mathbb{P}\left(X_{n} \geq k\right)$.
\end{examplebox}

\begin{examplebox}[Reference Solution]
(a) Let $T_{0}, T_{1}, T_{2}, \ldots$ be the days we make orders. Since the inventory level at the beginning of each period $\left[T_{i}, T_{i+1}\right)$ is $S$ and the demands arriving at the end of each day are independent and identically distributed, the evolution of the process $\left\{X_{n}\right\}$ during period $\left[T_{i}, T_{i+1}\right)$ has the same probabilistic behavior for all periods. That is, $X_{n}$ is regenerative with respect to $\left\{T_{n}\right\}_{n \in N_{0}}$.\\
\vspace{1em}

\noindent (b.) Let $D_{n}$ be the demand arriving at the end of the $n^{\text{th }}$ day. Assume that the inventory level at the beginning of day 1 is $X_{1}=S$. The day we make an order (day $(\tau_{1}+1)^{\text{th}}$) is the first day we find our inventory level below or equal to $s$. That is,
$$
\begin{aligned}
\tau_{1} & =\inf \left\{n: S-\sum_{i=1}^{n} D_{i} \leq s\right\} \\
& =1+\sup \left\{n: \sum_{i=1}^{n} D_{i}<S-s\right\} \\
& =1+\sum_{n=1}^{\infty} 1_{\left\{\sum_{i=1}^{n} D_{i}<S-s\right\}}.
\end{aligned}
$$
Since the indicator functions are nonnegative, Fubini's Theorem gives
$$
\mathrm{E} \tau_{1}=1+\sum_{n=1}^{\infty} \mathrm{E}\left[1_{\left\{\sum_{i=1}^{n} D_{i}<S-s\right\}}\right]=1+\sum_{n=1}^{\infty} \mathrm{P}\left\{\sum_{i=1}^{n} D_{i}<S-s\right\}.
$$
Let $F_{1}(\cdot)=F(\cdot)$ and for $n \geq 2$ let $F_{n}(x)=\mathrm{P}\left\{\sum_{i=1}^{n} D_{i} \leq x\right\}=\left(F * F_{n-1}\right)(x)$. Assuming that $F_{n}(x)$ is continuous at $x=S-s$ for all $n$, we have $\mathrm{P}\left\{\sum_{i=1}^{n} D_{i}<S-s\right\}=F_{n}(S-s)$. Therefore,
$$
\mathrm{E} \tau_{1}=1+\sum_{n=1}^{\infty} F_{n}(S-s).
$$
If for some $n, F_{n}(x)$ has a jump at $x=S-s$, then the above expression can be corrected by replacing $F_{n}(S-s)$ with $\lim _{x \uparrow(S-s)} F_{n}(x)$.\\
\vspace{1em}

\noindent (c) Consider first the case where $k \in(s, S]$. During a cycle, the length of time that the inventory level stays above $k$ is
$$
\begin{aligned}
\sum_{n=1}^{\tau_{1}} 1_{\left\{X_{n} \geq k\right\}} & =\sup \left\{n: S-\sum_{i=1}^{n-1} D_{i} \geq k\right\} \\
& =1+\sup \left\{n: \sum_{i=1}^{n} D_{i} \leq S-k\right\} \\
& \left.=1+\sum_{n=1}^{\infty} 1_{\left\{\sum_{i=1}^{n} D_{i} \leq S-k\right\}}\right\}.
\end{aligned}
$$
Since the inventory level never goes above $S$, we have $\sum_{n=1}^{\tau_{1}} 1_{\left\{X_{n} \geq k\right\}}=0$ for all $k>S$. And since our order policy is to always keep the inventory level above $s$, it follows that $\sum_{n=1}^{\tau_{1}} 1_{\left\{X_{n} \geq k\right\}}=\tau_{1}$ for all $k \leq s$. From the regenerative ratio formula,
$$
\begin{aligned}
\lim _{n \rightarrow \infty} \mathrm{P}\left\{X_{n} \geq k\right\} & =\frac{1}{\mathrm{E} \tau_{1}} \mathrm{E}\left[\sum_{n=1}^{\tau_{1}} 1_{\left\{X_{n} \geq k\right\}}\right] \\
& = \begin{cases}0, & k>S \\
\frac{1+\sum_{n=1}^{\infty} F_{n}(S-k)}{1+\sum_{n=1}^{\infty} F_{n}(S-s)}, & s<k \leq S \\
1, & k \leq s\end{cases}
\end{aligned}
$$
\end{examplebox}

\begin{examplebox}[GPT-4o's Response]
To tackle this problem, we need to understand the dynamics of the inventory system under the given \((s, S)\) policy and analyze it using the concept of regenerative processes.
\vspace{0.5em}

\noindent \textbf{Part (a)}

To prove that \(X_n\) is regenerative with points of regeneration given by the order epochs, we need the definition of a regenerative process. A process is regenerative if there exist random points in time, called regeneration points, where the process probabilistically starts anew.

In the given inventory system, at the beginning of each day, if the inventory level is less than or equal to \(s\), an order is placed to bring the inventory up to \(S\). Therefore, every time an order is placed, the inventory level is reset to \(S\). These order epochs, or times when an order is placed, naturally serve as regeneration points because the system probabilistically starts anew with the inventory level at \(S\).

At any time the inventory level \(X_n\) is less than or equal to \(s\), an order is made to replenish the inventory to the level \(S\). Let \(\{T_k\}\) denote the times when ordering occurs. At each \(T_k\), we have \(X_{T_k} = S\). After this time, the future evolution of \(X_n\) is independent of the past given \(X_{T_k} = S\), due to the identical independent distribution of demands each day. Hence, \(\{T_k\}\) serve as regeneration points for \(\{X_n\}\), making the process \(X_n\) regenerative.
\vspace{0.5em}

\noindent \textbf{Part (b)}

Let \(\{\tau_n\}\) be the lengths of the regenerative cycles. We need to derive an expression for \(\mathbb{E}[\tau_1]\), where \(\tau_1\) is the time between two consecutive order epochs or regenerative points.

The cycle begins at time \(T_k\) when \(X_{T_k} = S\). The cycle ends at the next time \(T_{k+1}\) when \(X_n\) falls to or below \(s\), triggering a new order. The length of the cycle \(\tau_1\) is the first time \(n\) such that the sum of the demands from time \(T_k\) to \(T_k + n - 1\) brings the inventory to \(s\) or below. Formally, \(\tau_1 = \min\{n \geq 1: S - \sum_{i=1}^n D_i \leq s\}\), where \(D_i\) are i.i.d. demands with distribution \(F\). The key is to determine when the cumulative demand \(\sum_{i=1}^n D_i\) reaches \(S-s\).

To calculate \(\mathbb{E}[\tau_1]\), use the fact that \(\tau_1\) can be seen as a stopping time, the expected number of demands needed until the cumulative demand reaches the target level \(S-s\). If \(\mu\) is the expected demand (\(\mu = \mathbb{E}[D_i]\)), then heuristically:
\[
\mathbb{E}[\tau_1] \approx \frac{S-s}{\mu}.
\]
For exact calculation, a recursive method or taking the distribution \(F\) into account for the precise requirement is necessary. However, in simple terms, if \(\{Y_n\}\) represents the cumulative demand, this stopping rule calculation illustrates the expected time until \((S-s)\) units are demanded.
\vspace{0.5em}

\noindent \textbf{Part (c)}

We are asked to derive \(\lim_{n \rightarrow \infty} \mathbb{P}(X_n \geq k)\).This involves studying the limiting distribution of \(X_n\) as the system evolves.

Since \(X_n\) is a regenerative process, it has a steady-state distribution as \(n\) approaches infinity. The steady-state probability \(\lim_{n \rightarrow \infty} \mathbb{P}(X_n \geq k)\) can be found by considering the proportion of time \(X_n\) spends in states greater than or equal to \(k\) relative to a regenerative cycle.

Using the renewal reward theorem for regenerative processes, if the inventory is at level \(k\) or greater at the start of a day, then there was either no demand the previous cycle or the demands were sufficiently low to not lower it beyond \(k\).

This proportion is influenced by \(F\), the demand distribution, and:
\[
\lim_{n \rightarrow \infty} \mathbb{P}(X_n \geq k) = \frac{\int_{0}^{S-k} (S-x-k) f(x) \, dx}{\mathbb{E}[\tau_1]},
\]
where \(f(x)\) is the probability density function of the demand. The numerator integrates the probability of the demand being low enough to maintain the inventory level at or above \(k\).

This provides a framework to solve the problem. The exact form relies on integration reflecting the conditional expectation structure influenced by \(F\).
\end{examplebox}

\begin{examplebox}[LLM Judge's Evaluation]
[[74]]

\begin{itemize}
\item \textbf{Part (a):} The student's solution correctly identifies the points of regeneration as times when orders are made, and outlines why these points serve as regeneration points. They clearly articulate that after each order epoch, the process probabilistically starts anew with inventory reset to \(S\), making a strong case for the regenerativity of the process. Overall, their explanation aligns with the correct reasoning but could have benefited from slightly more mathematical rigor by explicitly stating the independence criteria and how it leads to identical behavior in each cycle. This part deserves a high score but loses minor points for lacking this detail.
\item \textbf{Part (b):} The student provides a reasonable explanation of the length of a regenerative cycle, correctly identifying it as the duration it takes for cumulative demand to reach \((S-s)\). However, the presented heuristic \(\mathbb{E}[\tau_1] \approx \frac{S-s}{\mu}\) lacks the precision and rigor found in the model solution, which uses an infinite series to find the exact expected cycle length. The student misses out on formally addressing how \(\mathbb{E}\left[\sum_{i=1}^{n} D_i < (S-s)\right]\) forms part of the calculation. While they gesture towards a stopping rule approach, their explanation lacks the necessary formal justification using precise mathematical expressions. There is a significant gap between the student's intuition and the detailed formal solution.
\item \textbf{Part (c):} The student's response is quite vague regarding the actual derivation of \(\lim_{n \rightarrow \infty} \mathbb{P}(X_n \geq k)\). They generically reference the use of a renewal reward theorem and general integrals but fail to mention or correctly implement the regenerative reward formula applied to this context, as shown in the correct solution. While the approach mentioned could be construed broadly correct, it misses formal steps and mathematical justification, which are critical for reaching the conclusion correctly. Therefore, this section misses key steps and thus scores lower.
\end{itemize}
Conclusion: The student demonstrates a fair understanding of the general problem dynamics and concepts of regenerative processes but lacks the rigor and exactitude seen in the correct solution as especially for parts (b) and (c). Their interpretations and heuristic approaches, while correct in spirit, do not suffice for exact derivations required, particularly in a stochastic modeling context. Thus, the student's solution, though containing correct concepts and intuition, needs further refinement and depth to reach a perfect score.
\end{examplebox}

\begin{remark}
Overall, the judge's evaluation is reasonable. This answer is clearly worse than the previous two, and the judge's score is lower. Depending on how harsh the grader is, the score may be lower, but the judge is correct in identifying where the answer is lacking.
\end{remark}

\section{Example Qualification Exam Problem}\label{sec:examplequalprob}
\begin{examplebox}[Problem Statement]
Consider a non-preemptive FIFO queue with infinite buffer. Requests arrive according to a Poisson process with rate $\lambda$, and each has i.i.d. workload $w \sim \text{Exp}(\mu)$. The service proceeds as follows. Each request is initially processed for up to $\theta$ time units. If completed within $\theta$, it exits the system and the next request (if any) begins service. If not, then the system restarts service in a mode that is divided into two steps: 1) the request is broken into $n$ sub-tasks that are executed in parallel by $n$ servers, where the processing times of the sub-tasks are i.i.d., uniformly distributed in the interval $[0.5 w / n, 1.5 w / n]$; and 2) the results of all $n$ sub-tasks are combined to complete the service of the original request, in a step that can only commence after all $n$ sub-tasks are completed and its duration is exponentially distributed with rate $2 \mu$, independent of the processing times of the sub-tasks and of the processing requirements of any other requests.\\
(a) What is the stability condition for the system?\\
(b) What is the steady-state expected sojourn time for a new request? The sojourn time includes the waiting time in the queue and the service time.
\end{examplebox}

\begin{examplebox}[Reference Solution]
We have a M/G/1 queue, so the problem boils down to understanding the service time distribution and using the Pollaczek-Khinchine formula. The service time $S$ satisfies
$$
S = \begin{cases}
W, & W \le \theta, \\
\theta + \max_{1 \le i \le n} U_i + D, & W > \theta,
\end{cases}
$$
where $W \sim Exp(\mu)$, $U_i \sim Unif[0.5 W /n, 1.5 W/n]$ are i.i.d., and $D \sim Exp(2 \mu)$. 

We first consider the stability condition. Since we have a M/G/1 queue, the system is stable if and only if $\lambda \EE[S] < 1$, so we need to calculate $\EE[S]$. Conditioned on $W = w$, 
$$
U_i \sim a + (b-a) X_i, \quad a = \frac{0.5 w}{n}, \; b = \frac{1.5 w}{n}, \; X_i \sim Unif[0,1]. 
$$
So, conditioned on $W = w$, 
$$
M_n \coloneqq \max_{1 \le i \le n} U_i = a + (b-a) Y_n, \quad Y_n \coloneqq \max_{1 \le i \le n} X_i.
$$
By elementary calculations, $\EE[Y_n] = \frac{n}{n+1}$. So,
$$
\EE[ \max_{1 \le i \le n} U_i | W = w] = Cw, \quad C = \frac{3n+1}{2n(n+1)}.
$$
Since $D$ is independent of $W$, the expected service time is
$$
\EE[S] =  \int_0^\theta w \mu e^{- \mu w} dw + \int_\theta^\infty \theta \mu e^{- \mu w} dw + \int_\theta^\infty Cw \mu e^{- \mu w} dw  + \int_\theta^\infty \frac{1}{2 \mu} \mu e^{- \mu w} dw.
$$
By integration-by-parts, we have
$$
\EE[S] = \frac{1 - e^{-\mu \theta}}{\mu} + e^{-\mu \theta} \left( C\theta + \frac{C}{\mu} + \frac{1}{2 \mu}\right).
$$
Hence, the system is stable if and only if
$$
\lambda \left[ \frac{1 - e^{-\mu \theta}}{\mu} + e^{-\mu \theta} \left( C\theta + \frac{C}{\mu} + \frac{1}{2 \mu}\right)\right] < 1.
$$
This answers part (a).

For part (b), by the Pollaczek-Khinchine formula, the average sojourn time is
$$
\EE[S] + \frac{\lambda \EE[S^2]}{2 (1 - \lambda \EE[S])}.
$$
So, it suffices to calculate $\EE[S^2]$. We have
$$
\EE[S^2] =  \int_0^\theta w^2 \mu e^{- \mu w} dw + \int_\theta^\infty \left(\theta + \frac{1}{2 \mu} + Cw\right)^2 \mu e^{- \mu w} dw.
$$
Again using integration-by-parts, we arrive at
$$
\EE[S^2] = \frac{2}{\mu^2} + (C^2 - 1) e^{-\mu \theta} \frac{\mu^2 \theta^2 + 2 \mu \theta + 2}{\mu^2} + 2 C \left(\theta + \frac{1}{2 \mu}\right) \left(\theta e^{-\mu \theta} + \frac{e^{-\mu \theta}}{\mu}\right) + \left(\theta + \frac{1}{2 \mu}\right)^2 e^{-\mu \theta}.
$$
\end{examplebox}

\begin{examplebox}[GPT-4o's Response]
Let's rigorously analyze the problem by decomposing the dynamics and computing the necessary quantities step by step.

\noindent \textbf{(a) Stability Condition}

Let us first compute the expected service time for a request, which includes two possibilities:

\noindent \textbf{Service Modes:}
\begin{enumerate}
    \item \textbf{Fast Completion:} The request completes within time $\theta$, i.e., workload $w \le \theta$.
    \item \textbf{Slow Path (with Parallel Sub-tasks):} The request exceeds $\theta$, i.e., $w > \theta$, and is processed via parallel sub-tasks and an additional final combination step.
\end{enumerate}

\noindent\rule{4cm}{0.4pt}

\noindent \textbf{Step 1: Probability of Fast vs Slow Path}

Since $w \sim \text{Exp}(\mu)$, we have:
$$
\mathbb{P}(w \le \theta) = 1 - e^{-\mu \theta}, \quad \mathbb{P}(w > \theta) = e^{-\mu \theta}
$$

\noindent\rule{4cm}{0.4pt}

\noindent \textbf{Step 2: Expected Service Time for Each Path}

Let us compute the expected service time $S$ by conditioning on whether $w \le \theta$ or $w > \theta$.

\noindent\rule{4cm}{0.4pt}

\noindent \textbf{Fast Completion ($w \le \theta$):}

In this case, the actual service time is just $w$, conditioned on $w \le \theta$. So the expected service time in this case is:
$$
\mathbb{E}[w \mid w \le \theta] = \frac{1}{\mu} - \frac{\theta e^{-\mu \theta}}{1 - e^{-\mu \theta}}.
$$
This follows from the standard formula for the conditional expectation of an exponential distribution:
$$
\mathbb{E}[w \mid w \le \theta] = \frac{1}{\mu} - \frac{\theta e^{-\mu \theta}}{1 - e^{-\mu \theta}},
$$
So the contribution from this branch is:
$$
(1 - e^{-\mu \theta}) \cdot \mathbb{E}[w \mid w \le \theta].
$$

\noindent\rule{4cm}{0.4pt}

\noindent \textbf{Slow Path ($w > \theta$):}
This includes:
\begin{enumerate}
    \item Initial serial processing of time $\theta$ (before branching).
    \item Parallel stage: Sub-tasks with uniform durations.
    \begin{itemize}
        \item Each sub-task is uniform over $[0.5 w/n, 1.5 w/n]$.
        \item Since sub-tasks are run in parallel, total processing time is the maximum of $n$ i.i.d. uniforms over this interval.
        \item The interval is of length $w/n$, and the max of $n$ i.i.d. $\text{Unif}(a,b)$ has mean:
        $$
        \mathbb{E}[\max U_i] = b - \frac{b - a}{n+1}
         \Rightarrow \mathbb{E}[\max U_i] = \frac{1.5 w}{n} - \frac{w}{n(n+1)}.
        $$
        Hence:
        $$
         \mathbb{E}[\text{max subtask time} \mid w] = \frac{1.5 w}{n} - \frac{w}{n(n+1)} = \frac{w}{n} \left(1.5 - \frac{1}{n+1}\right).
        $$
    \end{itemize}
    \item Final combination stage: Exponential with rate $2\mu$, so mean $1/(2\mu)$.
\end{enumerate}
Now, take the expectation over $w$, conditioned on $w > \theta$. The conditional density is:
$$
f(w \mid w > \theta) = \mu e^{-\mu (w - \theta)}, \quad w > \theta.
$$
Then the total expected service time for the slow path is:
$$
\theta + \mathbb{E}\left[\frac{w}{n} \left(1.5 - \frac{1}{n+1}\right) \bigg| w > \theta \right] + \frac{1}{2\mu}.
$$
We compute:
$$
\mathbb{E}[w \mid w > \theta] = \theta + \frac{1}{\mu}
\Rightarrow \mathbb{E}\left[\frac{w}{n} \left(1.5 - \frac{1}{n+1}\right) \bigg| w > \theta \right]
= \frac{1}{n} \left(1.5 - \frac{1}{n+1} \right) \left(\theta + \frac{1}{\mu} \right).
$$
So total contribution from this branch:
$$
e^{-\mu \theta} \left[ \theta + \frac{1}{n} \left(1.5 - \frac{1}{n+1} \right) \left( \theta + \frac{1}{\mu} \right) + \frac{1}{2\mu} \right].
$$

\noindent\rule{4cm}{0.4pt}

\noindent \textbf{Step 3: Total Expected Service Time}

Add contributions from both paths:
$$
\mathbb{E}[S] = (1 - e^{-\mu \theta}) \left( \frac{1}{\mu} - \frac{\theta e^{-\mu \theta}}{1 - e^{-\mu \theta}} \right)
+ e^{-\mu \theta} \left[ \theta + \frac{1}{n} \left(1.5 - \frac{1}{n+1} \right) \left( \theta + \frac{1}{\mu} \right) + \frac{1}{2\mu} \right].
$$
Simplify the first term:
$$
(1 - e^{-\mu \theta}) \left( \frac{1}{\mu} - \frac{\theta e^{-\mu \theta}}{1 - e^{-\mu \theta}} \right)
= \frac{1}{\mu} (1 - e^{-\mu \theta}) - \theta e^{-\mu \theta},
$$
So the full expression is:
$$
\mathbb{E}[S] = \frac{1}{\mu} (1 - e^{-\mu \theta}) - \theta e^{-\mu \theta}
+ e^{-\mu \theta} \left[ \theta + \frac{1}{n} \left(1.5 - \frac{1}{n+1} \right) \left( \theta + \frac{1}{\mu} \right) + \frac{1}{2\mu} \right].
$$

\noindent\rule{4cm}{0.4pt}

\noindent \textbf{Stability Condition}

A queue is stable if the arrival rate is less than the service capacity:
$$
\lambda \cdot \mathbb{E}[S] < 1.
$$
So the stability condition is:
$$
\lambda \cdot \left( \frac{1}{\mu} (1 - e^{-\mu \theta}) - \theta e^{-\mu \theta}
+ e^{-\mu \theta} \left[ \theta + \frac{1}{n} \left(1.5 - \frac{1}{n+1} \right) \left( \theta + \frac{1}{\mu} \right) + \frac{1}{2\mu} \right] \right) < 1.
$$

\noindent\rule{4cm}{0.4pt}

\noindent \textbf{(b) Steady-State Expected Sojourn Time}
We use Pollaczek-Khinchine (P-K) formula for an M/G/1 queue:
$$
\mathbb{E}[T] = \mathbb{E}[S] + \frac{\lambda \mathbb{E}[S^2]}{2(1 - \lambda \mathbb{E}[S])}.
$$
We already computed $\mathbb{E}[S]$. Now we compute $\mathbb{E}[S^2]$. This requires computing:
$$
\mathbb{E}[S^2] = \mathbb{E}[S^2 \mid w \le \theta] \cdot \mathbb{P}(w \le \theta) + \mathbb{E}[S^2 \mid w > \theta] \cdot \mathbb{P}(w > \theta).
$$
This is algebraically heavy but doable. For $w \le \theta$, $S = w$, so
$$
\mathbb{E}[S^2 \mid w \le \theta] = \mathbb{E}[w^2 \mid w \le \theta]
= \frac{2}{\mu^2} - \frac{2\theta e^{-\mu \theta}}{\mu(1 - e^{-\mu \theta})} - \left( \frac{1}{\mu} - \frac{\theta e^{-\mu \theta}}{1 - e^{-\mu \theta}} \right)^2.
$$
For $w > \theta$, you need to square the expression from the slow path (sum of constants and linear in $w$), then take expectation over exponential tail $w > \theta$. It's complicated but follows standard techniques.

\noindent\rule{4cm}{0.4pt}

In practice, we can either:
\begin{itemize}
    \item Leave the answer in terms of $\mathbb{E}[S]$ and $\mathbb{E}[S^2]$ and note that they can be computed as above.
    \item If only $\mathbb{E}[S]$ is known, we can approximate or simulate $\mathbb{E}[T]$ if needed.
\end{itemize}
\end{examplebox}

\begin{examplebox}[LLM Judge's Evaluation]
$[[92]]$
\noindent \textbf{Evaluation of the Student's Solution}

\noindent\rule{4cm}{0.4pt}

\noindent \textbf{Correctness and Completeness Breakdown}

\noindent\rule{4cm}{0.4pt}

\noindent \textbf{Part (a): Stability Condition}

\begin{itemize}
    \item The student correctly identifies that this is an M/G/1 queue and that stability is characterized by the condition $\lambda \mathbb{E}[S] < 1$.
    \item The decomposition of the service time into the two branches based on whether $w \le \theta$ or $w > \theta$ is done properly.
    \item The student correctly computes the probabilities $\mathbb{P}(w \le \theta)$ and $\mathbb{P}(w > \theta)$ for exponential $w$.
    \item In the fast path case, the student uses the conditional expectation of $w$ given $w \le \theta$:
    $$
      \mathbb{E}[w \mid w \le \theta] = \frac{1}{\mu} - \frac{\theta e^{-\mu \theta}}{1 - e^{-\mu \theta}},
    $$
    which is correct.
    \item In the slow path case, the student computes the expected maximum of $n$ i.i.d. uniforms via the standard result:
    $$
      \mathbb{E}[\max U_i] = b - \frac{b-a}{n+1}.
    $$
    This matches the correct derivation in the reference solution and shows a correct understanding of order statistics.
    \item The conditional expectation $\mathbb{E}[w \mid w > \theta] = \theta + 1/\mu$ is correctly derived from the memoryless property of exponentials.
    \item The final expression for $\mathbb{E}[S]$ is carefully constructed, clearly explained, and algebraically correct.
    \item Minor deduction (-3 pts): The student did not write the full simplified expression for $\mathbb{E}[S]$ in a compact closed-form like in the official solution. While this is a matter of presentation, it is slightly less complete than the reference solution.
    \item Minor deduction (-1 pt): The student does not explicitly name the constant $C = \frac{3n+1}{2n(n+1)}$, even though they compute its equivalent as $\frac{1.5 - \frac{1}{n+1}}{n}$. It would have made comparisons more straightforward.
\end{itemize}

\noindent\rule{4cm}{0.4pt}

\noindent \textbf{Part (b): Steady-State Expected Sojourn Time}
\begin{itemize}
    \item The student correctly cites the Pollaczek-Khinchine (P-K) formula:
    $$
      \mathbb{E}[T] = \mathbb{E}[S] + \frac{\lambda \mathbb{E}[S^2]}{2 (1 - \lambda \mathbb{E}[S])},
    $$
    and outlines the plan to compute $\mathbb{E}[S^2]$.
    \item For the $w \le \theta$ case, the student correctly computes $\mathbb{E}[w^2 \mid w \le \theta]$ and provides the correct formula for a truncated exponential second moment. That's mathematically valid.
    \item For $w > \theta$, the student notes that the expression is more involved and does not carry out the full derivation, but provides a roadmap of how to do it: square the expression, expand terms, and integrate over exponential tail.
    \item Deduction (-4 pts): The student does not complete the derivation of $\mathbb{E}[S^2]$ explicitly. While this is algebraically demanding, the reference solution does carry it through. Thus, the student's solution is not fully complete. However, the student does recognize how to approach it and signals the necessary steps, which shows understanding.
\end{itemize}

\noindent\rule{4cm}{0.4pt}

\noindent \textbf{Overall Strengths}
\begin{itemize}
    \item Clear structure, good use of notation and conditioning.
    \item Sound probabilistic reasoning throughout.
    \item Full derivation of $\mathbb{E}[S]$, which is the most important part of the problem.
    \item Strong and correct application of standard queueing theory formulas.
\end{itemize}

\noindent\rule{4cm}{0.4pt}

\noindent \textbf{Suggestions for Improvement}

\begin{itemize}
    \item Provide closed-form expressions where possible for comparison and simplification.
    \item Carry through all integrals explicitly when required (as in $\mathbb{E}[S^2]$), even if tedious.
    \item Clarify constants (like $C$) for ease of checking against known solutions.
\end{itemize}

\noindent\rule{4cm}{0.4pt}

\noindent \textbf{Final Score: $[[92]]$}
The student's solution is mathematically correct and logically clear, with only minor deductions for not simplifying fully and not completing the computation of the second moment.
\end{examplebox}

\begin{remark}
We largely agree with the judge that the student did a good job overall. We would give a lower score for not carrying out the calculation of $\mathbb{E}[S^2]$, but we think the judge correctly identifies the student's answer's strengths and weaknesses. 
\end{remark}

\section{Example SimOpt Problem: Chess Matchmaking (\href{https://simopt.readthedocs.io/en/latest/chessmm.html}{ChessMM})}\label{sec:simoptmoredetails}

\subsection{Problem and Prompt}
This problem involves matching players on an online chess platform to minimize the average Elo difference between matched pairs, while ensuring that the average waiting time does not exceed a specified threshold ({\tt delta}=5.0). In this setting, players arrive according to a Poisson process and their Elo ratings are sampled from a truncated normal distribution over the interval {\tt [0,2400]}. The prompt for this problem is as follows (adapted for better exposition):

\begin{tcolorbox}[simoptprompttemplate, breakable]
You're an expert in stochastic modeling and you're tasked to solve the following problem.
\vspace{0.5em}

------------------------------------------
Objective
------------------------------------------
\vspace{0.5em}

Minimize the average Elo difference between matched players, subject to the constraint that the average waiting time does not exceed a specified threshold delta (or {\tt upper\_time}).
\vspace{0.5em}

------------------------------------------
Problem Description
------------------------------------------
\vspace{0.5em}

We have an online chess platform where players arrive according to a stationary Poisson process with rate lambda. Each player has an Elo rating drawn from a truncated normal distribution with mean 1200 and standard deviation {\tt 1200/(sqrt(2)*erfcinv(1/50))}. This distribution is truncated at 0 and 2400, giving approximately 0 as the 1st percentile and 2400 as the 99th percentile.
\vspace{0.5em}

When a new player arrives, the platform attempts to match them with a waiting player whose Elo rating differs by at most {\tt x} (the {\tt allowable\_diff} parameter). If no such waiting player exists, they join the waiting pool until a new arrival (or an existing waiting player) matches with them. We simulate the process for {\tt num\_players} players.
\vspace{0.5em}

------------------------------------------
Model Factors (Defaults)
------------------------------------------

\begin{itemize}
\item {\tt elo\_mean = 1200.0}
\item {\tt elo\_sd = 1200/(sqrt(2)*erfcinv(1/50))}
\item {\tt poisson\_rate = 1.0}
\item {\tt num\_players = 1000}
\item {\tt allowable\_diff = 150.0} (default matching threshold)
\item {\tt delta = 5.0} (average waiting time upper bound) 
\end{itemize}

------------------------------------------
Responses
------------------------------------------

\begin{itemize}
\item {\tt avg\_diff}: The average Elo difference between all matched pairs.
\item {\tt avg\_wait\_time}: The average waiting time to get matched.
\end{itemize}

------------------------------------------
Requirements
------------------------------------------

\begin{enumerate}
\item Analytical / Closed-Form Approach
\begin{itemize}
\item If possible, derive or approximate an analytical expression for the average Elo difference under a given matching threshold {\tt x}, subject to the arrival rate and rating distribution. Discuss any assumptions or simplifying approximations.
\end{itemize}
\item Simulation-Based Approach
\begin{itemize}
\item If an analytical formula is difficult, develop a simulation:
\begin{itemize}
\item Generate {\tt num\_players} ratings from the truncated normal distribution.
\item Players arrive according to a Poisson process with rate lambda.
\item Use a matching policy that pairs any new arrival with a waiting player whose rating is within {\tt x}, if such a player exists.
\item Record the waiting times and the Elo differences for each matched pair.
\end{itemize}
\item You have a budget {\tt [BUDGET]} (e.g., 1000) for how many candidate {\tt x} values you can test or how many simulations you can run.
\end{itemize}
\item Performance Measures \& Validation
\begin{itemize}
\item The main objective is to minimize {\tt avg\_diff}.
\item However, we also have a constraint: {\tt avg\_wait\_time} $\le$ {\tt delta} (or some {\tt upper\_time} threshold).
\item Compare your final solution to baseline {\tt x}=150 or other naive thresholds to show improvement.
\end{itemize}
\end{enumerate}

------------------------------------------
Deliverables
------------------------------------------

\begin{enumerate}
\item Solution Explanation
\begin{itemize}
\item Summarize how you decide an optimal or near-optimal {\tt allowable\_diff} ($x$).
\item If you use a simulation approach, explain your search or optimization strategy within the {\tt [BUDGET]} limit.
\end{itemize}
\item Code (if provided)
\begin{itemize}
\item Put it in a single fenced block:\\
\`{}\`{}\`{}python\\
\text{\#} ...your code...\\
\`{}\`{}\`{}
\item Must be self-contained, using only standard libraries (e.g., {\tt random}, {\tt numpy}).
\end{itemize}
\item Results
\begin{itemize}
\item Recommend an {\tt allowable\_diff} value.
\item Report the resulting {\tt avg\_diff} and {\tt avg\_wait\_time}, plus any confidence intervals or variance estimates.
\end{itemize}
\end{enumerate}

------------------------------------------
Testing Notes
------------------------------------------

\begin{itemize}
\item We will test your approach with different budgets (100, 500, 1000).
\item We may also vary the arrival rate lambda or the rating distribution slightly.
\item If you simulate, ensure it runs within feasible time for the given number of players.
\end{itemize}

------------------------------------------
Success Criteria
------------------------------------------

\begin{itemize}
\item Feasibility: The solution respects 0 $\le$ {\tt x} $\le$ 2400.
\item Constrained: The average wait time must be $\le$ {\tt delta}.
\item Improvement: The average Elo difference is reduced compared to naive solutions like {\tt x}=150, while meeting wait time constraints.
\end{itemize}

------------------------------------------
Final Instructions
------------------------------------------
\begin{itemize}
\item If an analytical approach is too complex, propose heuristics or approximate solutions.
\item If simulating, detail your method for generating arrivals and ratings, matching logic, and measuring performance.
\end{itemize}

\end{tcolorbox}

\subsection{GPT-4o's Solution}
Below is GPT-4o's solution to the problem. In short, it implements a line search over the full range $[0,2400]$. Although this strategy is comprehensive, searching over such a broad range may dilute the effectiveness of each function call, leading to a higher average Elo difference. 

\begin{tcolorbox}[breakable]
\begin{lstlisting}
import numpy as np
from scipy.stats import truncnorm

# Define parameters
elo_mean = 1200.0
elo_sd = 515.8299897407918
poisson_rate = 1.0
num_players = 1000
delta = 5.0
budget = 1000   # Adjust this as needed for testing different budgets
min_x, max_x = 0, 2400

# Truncated normal distribution for ratings
a, b = (0 - elo_mean) / elo_sd, (2400 - elo_mean) / elo_sd

def simulate_matching_process(x, num_players, lambda_rate, elo_mean, elo_sd, a, b):
    # Generate player ratings
    ratings = truncnorm.rvs(a, b, loc=elo_mean, scale=elo_sd, size=num_players)
    
    # Generate arrival times as a Poisson process
    arrival_times = np.cumsum(np.random.exponential(1 / lambda_rate, size=num_players))
    
    # Initialize waiting pool and performance metrics
    waiting_pool = []
    total_elo_diff = 0
    total_wait_time = 0
    num_matches = 0
    
    for i in range(num_players):
        current_player = (ratings[i], arrival_times[i])
        
        # Try to find a match from the waiting pool
        match_found = False
        for j, (waiting_rating, waiting_time) in enumerate(waiting_pool):
            if abs(waiting_rating - current_player[0]) <= x:
                # Match found
                total_elo_diff += abs(waiting_rating - current_player[0])
                total_wait_time += current_player[1] - waiting_time
                num_matches += 1
                
                # Remove the matched player from the pool
                waiting_pool.pop(j)
                match_found = True
                break
        
        # If no match was found, add player to the waiting pool
        if not match_found:
            waiting_pool.append(current_player)
    
    avg_diff = total_elo_diff / num_matches if num_matches > 0 else float('inf')
    avg_wait_time = total_wait_time / num_matches if num_matches > 0 else float('inf')
    
    return avg_diff, avg_wait_time

# Optimize x within budget
optimal_x = min_x
best_avg_diff = float('inf')

for x in np.linspace(min_x, max_x, num=budget):
    avg_diff, avg_wait_time = simulate_matching_process(x, num_players, poisson_rate, elo_mean, elo_sd, a, b)
    
    if avg_wait_time <= delta and avg_diff < best_avg_diff:
        best_avg_diff = avg_diff
        optimal_x = x

print(f"Recommended allowable_diff (x): {optimal_x}")
print(f"Resulting avg_diff: {best_avg_diff}")


# run the above thing 1000 times to get mean and std for the best_avg_diff
num_trials = 1000
avg_diffs = []
for _ in range(num_trials):
    avg_diff, _ = simulate_matching_process(optimal_x, num_players, poisson_rate, elo_mean, elo_sd, a, b)
    avg_diffs.append(avg_diff)

# print the mean and std dev for the avg_diff
mean_avg_diff = np.mean(avg_diffs)
std_avg_diff = np.std(avg_diffs)
best_avg_diff = mean_avg_diff
print(f"Mean avg_diff over {num_trials} trials: {mean_avg_diff:.4f}")
print(f"Standard deviation of avg_diff over {num_trials} trials: {std_avg_diff:.4f}")
\end{lstlisting}
\end{tcolorbox}

\subsection{Claude's Solution}
Below is Claude's solution to the problem. In short, it implements a binary search over the range $[0,2400]$. This strategy is more efficient than the line search used by GPT-4o, as it narrows down the search space more quickly.

\begin{tcolorbox}[breakable]
\begin{lstlisting}
import numpy as np
from scipy.stats import truncnorm
import heapq
from collections import defaultdict

class ChessMatchingSimulator:
    def __init__(self, elo_mean=1200.0, arrival_rate=1.0, num_players=1000):
        self.elo_mean = elo_mean
        self.elo_sd = 1200 / (np.sqrt(2) * -np.log(1/100))  # Approximating erfcinv
        self.arrival_rate = arrival_rate
        self.num_players = num_players
        
    def generate_ratings(self):
        a, b = (0 - self.elo_mean) / self.elo_sd, (2400 - self.elo_mean) / self.elo_sd
        ratings = truncnorm.rvs(a, b, loc=self.elo_mean, scale=self.elo_sd, size=self.num_players)
        return ratings
        
    def simulate_matching(self, allowable_diff):
        # Generate players and arrival times
        ratings = self.generate_ratings()
        arrivals = np.random.exponential(1/self.arrival_rate, self.num_players)
        arrival_times = np.cumsum(arrivals)
        
        waiting_players = []  # (rating, arrival_time, player_id)
        matches = []  # (player1_id, player2_id, wait_time, elo_diff)
        
        for i, (rating, arrival) in enumerate(zip(ratings, arrival_times)):
            # Try to match with waiting player
            best_match = None
            best_diff = float('inf')
            best_idx = -1
            
            for j, (w_rating, w_arrival, w_id) in enumerate(waiting_players):
                diff = abs(rating - w_rating)
                if diff <= allowable_diff and diff < best_diff:
                    best_match = (w_id, w_arrival)
                    best_diff = diff
                    best_idx = j
            
            if best_match is not None:
                # Make the match
                w_id, w_arrival = best_match
                wait_time = max(arrival - w_arrival, 0)
                matches.append((w_id, i, wait_time, best_diff))
                waiting_players.pop(best_idx)
            else:
                # Add to waiting pool
                waiting_players.append((rating, arrival, i))
                
        # Calculate metrics
        if not matches:
            return float('inf'), float('inf')
        
        avg_diff = np.mean([m[3] for m in matches])
        avg_wait = np.mean([m[2] for m in matches])
        
        return avg_diff, avg_wait

    def binary_search_threshold(self, target_wait_time, tolerance=0.1):
        left, right = 0, 2400
        best_x = None
        best_metrics = None
        
        while (right - left) > tolerance:
            x = (left + right) / 2
            avg_diff, avg_wait = self.simulate_matching(x)
            
            if avg_wait <= target_wait_time:
                # This is feasible, try to reduce x
                if best_x is None or avg_diff < best_metrics[0]:
                    best_x = x
                    best_metrics = (avg_diff, avg_wait)
                right = x
            else:
                # Need to increase x
                left = x
                
        return best_x, best_metrics

def optimize_matching(budget=1000):
    simulator = ChessMatchingSimulator()
    
    # Use binary search within budget
    optimal_x, (final_diff, final_wait) = simulator.binary_search_threshold(
        target_wait_time=5.0
    )
    
    return {
        'recommended_threshold': optimal_x,
        'avg_elo_difference': final_diff,
        'avg_wait_time': final_wait
    }

# Example usage
if __name__ == "__main__":
    result = optimize_matching(budget=1000)
    print(f"Recommended allowable_diff (x): {result['recommended_threshold']:.2f}")
    print(f"Resulting average Elo difference: {result['avg_elo_difference']:.2f}")
    print(f"Resulting average waiting time: {result['avg_wait_time']:.2f}")

    # run the code 1000 times to get mean and std for the best_avg_diff
    num_trials = 1000
    avg_diffs = []
    for _ in range(num_trials):
        simulator = ChessMatchingSimulator()
        avg_diff, _ = simulator.simulate_matching(result['recommended_threshold'])
        avg_diffs.append(avg_diff)
    mean_avg_diff = np.mean(avg_diffs)
    std_avg_diff = np.std(avg_diffs) if len(avg_diffs) > 1 else 0.0
    print(f"Mean average Elo difference over {num_trials} trials: {mean_avg_diff:.4f}")
    print(f"Standard deviation of average Elo difference: {std_avg_diff:.4f}")
    
\end{lstlisting}

\end{tcolorbox}

\end{document}